\documentclass[10pt,twocolumn,letterpaper]{article}

\usepackage{cvpr}              
\usepackage[accsupp]{axessibility}
\usepackage{graphicx}
\usepackage{amsmath}
\usepackage{amssymb}
\usepackage{booktabs}
\usepackage{multirow}
\usepackage{comment}
\usepackage{color, colortbl}
\usepackage{adjustbox}
\usepackage{gensymb}
\usepackage{bbding}
\usepackage{amssymb}
\usepackage{pifont}
\usepackage{fixltx2e}
\usepackage{appendix}

\usepackage[pagebackref,breaklinks,colorlinks]{hyperref}

\usepackage[capitalize]{cleveref}
\crefname{section}{Sec.}{Secs.}
\Crefname{section}{Section}{Sections}
\Crefname{subsection}{Subsection}{Subsections}
\crefname{subsection}{Sec.}{Secs.}
\Crefname{table}{Table}{Tables}
\crefname{table}{Tab.}{Tabs.}
\Crefname{figure}{Figure}{Figures}
\crefname{figure}{Fig.}{Figs.}
\Crefname{equation}{Equation}{Equations}
\crefname{equation}{Eq.}{Eqs.}

\begin{document}
\title{DAD-3DHeads: A Large-scale Dense, Accurate and Diverse\\
Dataset for 3D Head Alignment from a Single Image} 



\author{
Tetiana Martyniuk\textsuperscript{1,2 *}\
Orest Kupyn\textsuperscript{1,2 *} \ 
Yana Kurlyak\textsuperscript{1,2} \ 
Igor Krashenyi\textsuperscript{1,2} \\  
Ji\v{r}i Matas\textsuperscript{3} \ 
Viktoriia Sharmanska\textsuperscript{4, 5} \ 
\\
\textsuperscript{1} Ukrainian Catholic University \ 
\textsuperscript{2} Piñata Farms, Los Angeles, USA \\
\textsuperscript{3} Visual Recognition Group, Center for Machine Perception, FEE, CTU in Prague \\
\textsuperscript{4} University of Sussex \ 
\textsuperscript{5} Imperial College London \\
}

\twocolumn[{%
\renewcommand\twocolumn[1][]{#1}%
\maketitle
\begin{center}
    \centering
    \captionsetup{type=figure}

    
    \includegraphics[width=\textwidth]{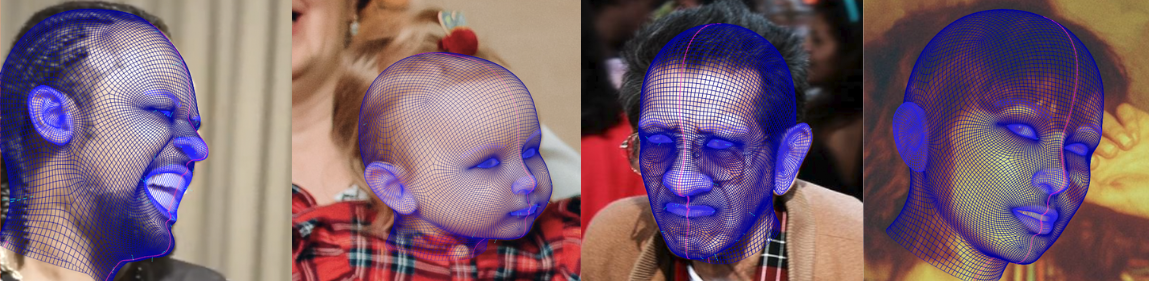}

    \caption{\textbf{DAD-3DHeads}, a \textbf{D}ense, \textbf{A}ccurate, and \textbf{D}iverse 3D Head dataset, labeled with over 3.5K verified accurate landmarks. 
    A model trained on DAD-3DHeads achieves superior performance on diverse 3D head tasks. It is robust to domain shifts common in the wild, including head pose changes, occlusions, facial expressions, age groups, illumination conditions, and image quality. 
    Best viewed in color.}
\end{center}
}
]
\maketitle

\def\thefootnote{*}\footnotetext{These authors contributed equally to this work.}

\begin{abstract}
We present DAD-3DHeads, a dense and diverse large-scale dataset, and a robust model for 3D Dense Head Alignment in-the-wild. 
It contains annotations of over 3.5K landmarks that accurately represent 3D head shape compared to the ground-truth scans.
The data-driven model, DAD-3DNet, trained on our dataset, learns shape, expression, and pose parameters, and performs 3D reconstruction of a FLAME mesh.
The model also incorporates a landmark prediction branch to take advantage of rich supervision and co-training of multiple related tasks.
Experimentally, DAD-3DNet outperforms or is comparable to the state-of-the-art models in (i) 3D Head Pose Estimation on AFLW2000-3D and BIWI, (ii) 3D Face Shape Reconstruction on NoW and Feng, and (iii) 3D Dense Head Alignment and 3D Landmarks Estimation on DAD-3DHeads dataset. 
Finally, diversity of DAD-3DHeads in camera angles, facial expressions, and occlusions enables a benchmark to study in-the-wild generalization and robustness to distribution shifts. 
The dataset webpage is \href{https://p.farm/research/dad-3dheads}{https://p.farm/research/dad-3dheads}.
\end{abstract}
\vspace{-0.5cm}
\section{Introduction}
\label{sec:intro}
Tremendous progress in 3D face analysis has been made since the first 3D morphable model (3DMM) \cite{3dmm} from an image had been proposed \cite{3dmm_survey}. 
The use cases for precise 3D face models are abundant: accurate face recognition and face detection \cite{FacePoseNet}, realistic 3D avatars and animation for VR and games \cite{LipSync3D}, face re-enactment and synthesis for dubbing \cite{NeuralVoicePuppetry}, virtual mirrors and try-on, statistical shape models for medical tasks such as segmentation and analysis of variations in anatomical structures \cite{zheng2017statistical}. 

These applications require not only accurate 3D face geometry but also (1) handling the diversity, e.g., ethnic, age, gender subgroups, and (2) generalizing to in-the-wild deployment conditions, i.e., beyond controlled capture and beyond the data they are trained on.  
The largest face models up-to-date \cite{Ploumpis_2019_CVPR, 7780967} have focused on the (1) aspect by collecting diverse 3D face and head scans, and building 3DMMs models for different age, gender and ethnicity. 
In-the-wild generalization has been identified as a pressing challenge of the next generation 3D face models \cite{3dmm_survey}. 
This (2) aspect of in-the-wild generalization is the focus of our study. 

The progress that we have witnessed with deep learning has impacted closely related facial analysis tasks such as Landmark Localisation \cite{Sun_2013_CVPR, HyperFace, sagonas2013300, BurgosArtizzu2013RobustFL, RetinaFace}, Facial Alignment in 2D and 3D \cite{DBLP:journals/corr/ZhangZL016, 10.1007/978-3-319-10599-4_8, img2pose, DAMDNet, bulat_how_far, 3DFAW, SADRNet, bulat_binarized, NME_bbox2, wayne2018lab}, and Face Detection \cite{HyperFace, DBLP:journals/corr/ZhangZL016, 10.1007/978-3-319-10599-4_8, img2pose, occlusion_coherence, RetinaFace}. 
This has been driven by the community effort towards collecting and annotating large image datasets captured in unconstrained conditions, building enhanced models that can take advantage of such large datasets, and most importantly \emph{openness}, i.e., making the models and datasets publicly available for research use. 

Nevertheless, 3D face or head alignment from a single image in the wild remains an open challenge. 
The difficulty comes from (1) lack of 2D-3D ground-truth data and, as a result, (2) ambiguity of the task and reliance on 3D shape priors. 
Many methods have been developed to fill the gap of missing 2D-3D annotations (1), primarily using 2D landmarks datasets for fitting, or exploring extra knowledge such as identity invariance \cite{RingNet}, or co-training with related face detection \cite{RetinaFace}, \cite{FacePoseNet} tasks to drive the recovery of 3D face geometry. 
Up until now, evaluation of the efficiency of these approaches has been problematic due to the lack of ground-truth data. 
Regarding (2), the state-of-the-art 3D face reconstruction methodologies such as non-linear 3DMMs and deep learning models \cite{KF-ITW, 7780967, LargeScale3DMM, FLAME, Ploumpis_2019_CVPR} are based on learning a statistical 3D facial model and fitting it to the image as a shape (or shape and texture) prior. 
This direction has a long history tracing back to the seminal work of Blanz and Vetter \cite{3dmm}. 
It relies on a large and diverse dataset of 3D/4D scans to build the statistical 3D face model that can be decomposed into facial shape (identity and expression), and the camera parameters. 
This comes at the cost of laborious data collection with expensive 3D acquisition devices, and the fact that 3D acquisition devices cannot operate in arbitrary conditions. 
Hence, the current 3D facial databases have limited data sample size and have been captured not-quite-in-the-wild \cite{RingNet}. 

In this work, we show that \emph{without expensive devices, like scanners, that are difficult to deploy in the wild, we can collect accurate annotations of 3D landmarks directly from images, which is labor-efficient and effective to push the state-of-the-art results for 3D head recovery from images}. 
Our contributions are as follows:
\begin{itemize}
\vspace{-0.5em}
    \item
    A new \textbf{D}ense, \textbf{A}ccurate and \textbf{D}iverse dataset for 3D Dense Head Alignment in-the-wild, \textbf{DAD-3DHeads}.
    It has over 3.5K verified accurate landmarks, the densest annotations for 3D dense head alignment in-the-wild currently available. DAD-3DHeads contains a variety of extreme poses, facial expressions, challenging illuminations, and severe occlusions cases.
    Accuracy and consistency of the annotations are compared to the ground truth 4D scans and head pose labels.
    \vspace{-0.5em}
    \item 
    A novel way to address the problems of shape reconstruction and pose estimation \emph{simultaneously} during training via optimizing two loss components: (i) \textbf{Shape+Expression Loss} and (ii) \textbf{Reprojection Loss}. 
    (i) is based on the normalized 3D vertices that enables disentangling the shape and expression information from the pose; (ii) is based on the full head dense 2D landmarks and assesses the pose accuracy.
    That makes the rich annotations fully utilized, which could not have been done previously due to the lack of GT annotations. 
    Extensive ablation studies show the importance of both loss components.
    \vspace{-0.5em}
    \item 
    \textbf{DAD-3DNet} model that maps an input image to 3D mesh representation consistent with the FLAME topology. 
    The model is trained end-to-end by regressing the 3DMM parameters and recovering the 3D head geometry with differential FLAME decoder. 
    The proposed approach learns the head shape, pose, and expression simultaneously. 
    DAD-3DNet outperforms state-of-the-art on a range of tasks, suggesting that dense supervision as provided in our dataset, enables a holistic framework for 3D Head Analysis from images. 
    \vspace{-0.5em}
    \item  
    A novel benchmark with the evaluation protocol for quantitative assessment of 3D dense head fitting, i.e. 3D Head Estimation from dense annotations.
    Our evaluation protocol introduces two novel metrics: 
    \textbf{Reprojection NME} computing the NME of the reprojected 3D vertices onto the image plane, and \textbf{\textit{Z\textsubscript{n}} Accuracy} evaluating the ordinal distance of the $Z$-coordinate and accuracy of the 3D fitting. 
\end{itemize}
\section{Related Work}

This section provides an overview of the available 3D face datasets, followed by a survey of the methods targeting 3D head-related tasks. 

{\bf 3D Face Datasets.} Existing 3D face datasets differ  based on registration of a 3D face model. Model fitting datasets \cite{rot_matrix_metrics, booth_2018, KF-ITW} fit the 3DMM to the images, which makes it suitable for large-scale datasets. The main limitation of such approach is shape detalization. To get a precise 3D facial shape, multi-view camera systems are applied \cite{BP4D, 6126510} or depth camera \cite{Bosphorus, BJUT, 4813324, 1613022, 7208488, cheng20174dfab}, however, these sensors suffer from limited spatial resolution. The FaceScape dataset \cite{FaceScape} contains textured 3D faces recorded using a dense camera array under controlled lighting, which retrieves the 3D facial model preserving low-level details such as small wrinkles and pores. 
The 3DFAW-Video dataset \cite{3DFAW-Video} lacks subjects diversity, and is not really "in-the-wild"; 300W-LP \cite{sagonas2016300, AFLW2000-3D} is synthetic and focuses only on faces.
In contrast to our dataset, none of the datasets is diverse, accurate, dense, and in-the-wild at the same time. 

\textbf{3D Head Pose Estimation}. 
\textit{Classical methods} for head pose estimation are based on traditional techniques such as cascade detectors \cite{Viola01rapidobject}  or template matching \cite{brunelli2009template}. 
Cascade detectors localize the head for each pose \cite{Kwong2002CompositeSV}, while a template matching approach compares query image with a set of pre-labeled templates and finds a corresponding pose \cite{HeadPoseEstimation, PoseDiscrimination}. \textit{Geometric methods} use facial landmarks retrieved from the input image and estimate the head pose empirically  \cite{Burger2014SelfinitializingHP, 557266}. \textit{Regression and classification methods} include wide-ranging methods that fit a mathematical model to predict the head pose from labeled training data or discretized set of poses\cite{1047456, head_pose_driver_monitoring, 7279167, nataniel, 6751256, benfold2008headpose}. \emph{Multi-task} approach combines a head pose estimation learning with other facial analysis tasks, such as Face Detection \cite{All-In-One, HyperFace, zhu2012face}, Face Recognition \cite{All-In-One}, Landmark Localization \cite{All-In-One, HyperFace, zhu2012face}, Alignment \cite{10.1007/978-3-319-10599-4_8, DBLP:journals/corr/ZhangZL016, All-In-One}. Our approach is related to the latter one, where the study of 3D Head Reconstruction is coupled with learning parameters of a 3D head model and Landmark Localization.

\textbf{3D Face Alignment}. Early 3D Morphable Face Models (3DMM) \cite{3dmm, bazelface} 
were derived from a small amount of registered 3D scans, e.g.,  Basel Face Model (BFM) \cite{bazelface} has 200 human faces. 
More recent models such as FLAME \cite{FLAME} are learnt from a significantly larger amount of scans, i.e., FLAME uses 3,800 3D scans of human heads. 
Nevertheless, the diversity of the scans is limited. 

RingNet\cite{RingNet} is trained to estimate the 3D face shape from a single image without direct 3D supervision to overcome this limitation. 
In contrast, we train our model to perform 3D head reconstruction from an image directly by the 2D-3D supervision as provided in our dataset.
Similarly motivated is 3DDFA \cite{zhu2017face}, a Cascaded CNN model which directly predicts a dense 3DMM from the facial image. 
This approach has been further extended and optimized in \cite{guo2020towards} with meta-joint optimization to facilitate parameters regression. 
Another approach called DECA \cite{DECA} is trained to regress a parameterized face model. 
A recent FAN model \cite{bulat_how_far} has been constricted by stacking four Hourglass models \cite{newell, StackedHourglass} in which all bottleneck blocks were replaced with the hierarchical, multi-scale and parallel binary residual blocks \cite{bulat_binarized}.
Instead of using Landmark Localization, in \cite{FacePoseNet} the authors propose to align human faces directly from an image using 6 degrees of freedom (6DoF 3D) -- rotations and translations along $X$, $Y$, $Z$ axis. 
\cite{DAMDNet} introduces a model with a lightweight attention mechanism for Face Alignment. 
In contrast, we collect a large-scale, diverse dataset with annotations directly in 3D and correspond with FLAME topology. This enables efficient training of the DAD-3DNet for a range of 3D head tasks. 
\section{DAD-3DHeads Dataset}

To create a large-scale dataset of in-the-wild images, we repurpose a modern 3D
modeling tool and introduce a novel annotation scheme that addresses the problems exhibited by existing labeling tools, such as "guessing" the positions of the correct landmarks for invisible parts of the head, thus enabling accurate annotations for any head images.
In this section we verify that obtained annotations are accurate compared to the GT 3D scans, and of high quality, i.e., reducing annotator's errors by half.

\subsection{Data acquisition}\label{ssec:labeling}

\begin{figure}[htp!]
\centering
\includegraphics[width=0.47\textwidth]{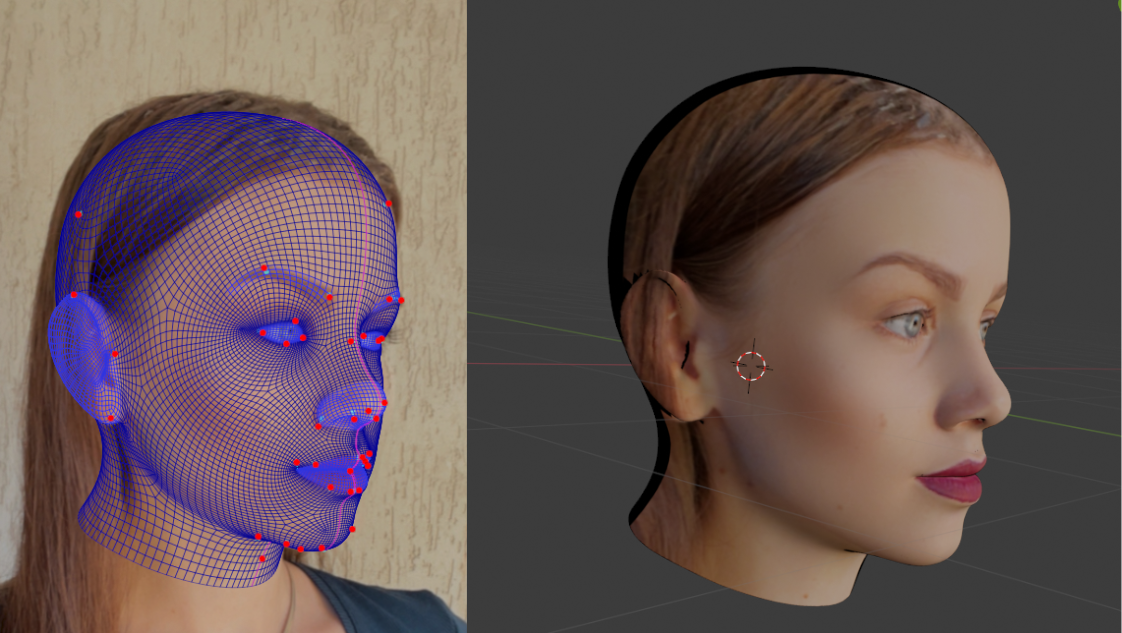}
  \caption{\textbf{A labeling tool example}. The annotator fits the 3D Head model to the image by anchoring pinpoints. 
  The corresponding 3D textured render is available to ensure the visual plausibility of the head shape.}
  \vspace{-1em}
  \label{fig:labeling_tool}
\end{figure}

We fit a 3D Morphable Model of a human head to a given photo with a simple interface. 
The annotators do not explicitly control or label either the 3DMM parameters or the blendshapes.
The fitting is conditioned upon the visible part of the head and the prior FLAME model \cite{FLAME}.
The annotators "pin" the points on the 3D mesh surface (see \cref{fig:labeling_tool}, left) to the specific pixels of the image.
The mesh then undergoes the optimization of the 3DMM parameters, so that the "pin" reprojection error is minimized.
During the labeling process, labelers can see the texture rendered onto the 3D mesh with respect to their fitting to verify that the results are visually plausible (\cref{fig:labeling_tool}, right). 
We use the 2D reprojection of the 3D mesh onto the image to ensure that the boundaries of the facial features and the skull are correct, and the relative depth information to confirm that the image provides realistic texture mapping onto the human head model.
The details of the annotation procedure along with the visuals - images of the intermediate steps and the full video example - are provided in \cref{ssec:annotation_process_detailed}.
In total we receive 5,023 dense landmarks consistent with the FLAME topology, namely, FLAME mesh vertices. 

%

\subsection{Dataset Statistics}\label{ssec:data_statistics}

\begin{figure}[t]
\includegraphics[width=0.4\textwidth]{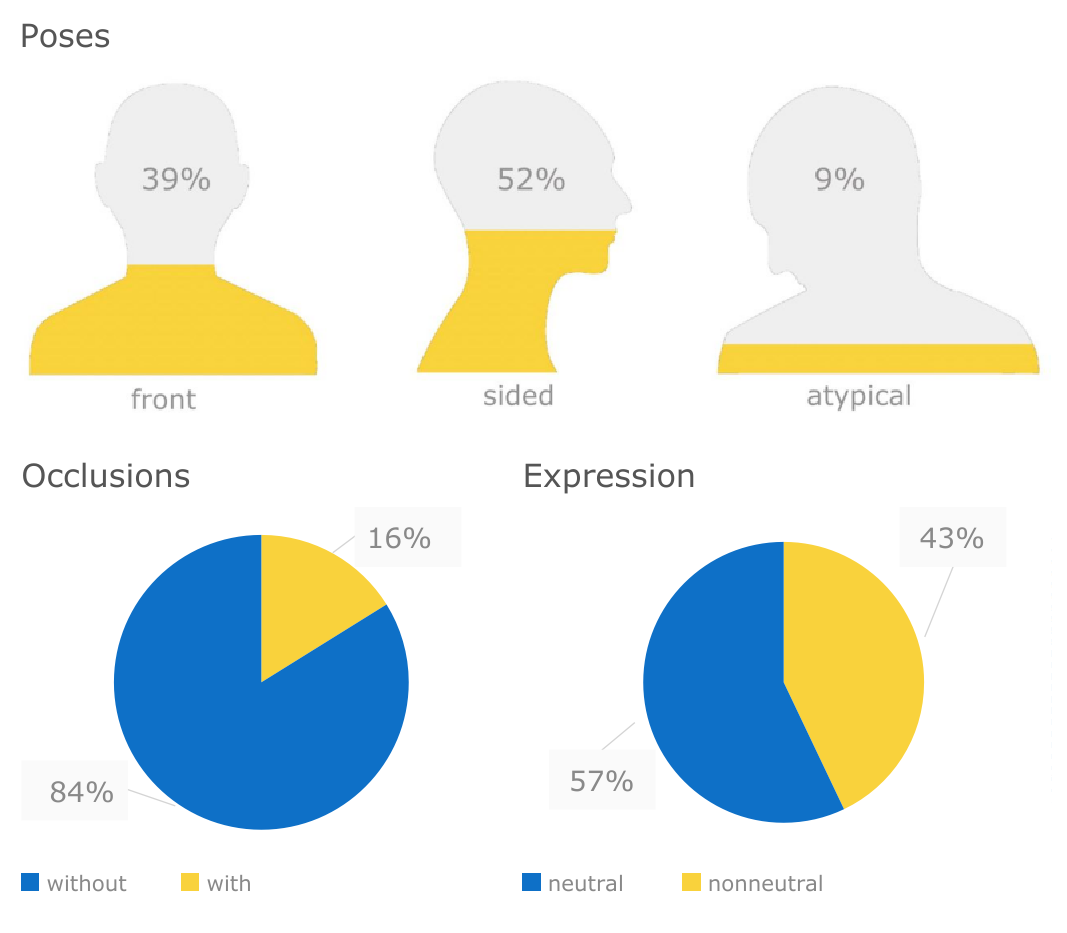}
\centering
\caption{\textbf{Dataset Properties:} DAD-3DHeads is well balanced over a wide range of poses, face expressions, and occlusions. The attribute labels are a valuable signal for subgroup analysis and for generalization to in-the-wild deployment conditions.}
\label{fig:dataset_stats}
\end{figure}

DAD-3DHeads dataset consists of 44,898 images collected from various sources (37,840 in the training set, 4,312 in the validation set, and 2,746 in the test set).
For each image, we provide 5,023 vertices of the FLAME mesh, 3,669 of which are accurately labeled (we demonstrate it in \cref{ssec:accuracy_labels}), neck and eyeballs excluded. 
We refer to this subset of 3,669 landmarks as "head" (see \cref{fig:head_and_face}).
We also provide the model-view and frustum projection matrices that map the 3D mesh from model space onto the image for different training scenarios and evaluation purposes. 
In addition, we release rich attribute information, showing the variability and unbiasedness of the data. 
DAD-3DHeads attributes include head poses, presence of emotions, occlusions (see \cref{fig:dataset_stats}), as well as gender, age group, image quality, and illumination labels. 
The detailed dataset card can be found in \cref{ssec:dataset_card_appendix}.

\subsection{Annotation accuracy}\label{ssec:accuracy_labels}


To check the accuracy of our annotations, we calculate the accuracy of the head shape reconstruction and head pose estimation compared to ground-truth 3D scans. 

\begin{figure*}[t!]
  \centering
  \includegraphics[width=0.82\textwidth]{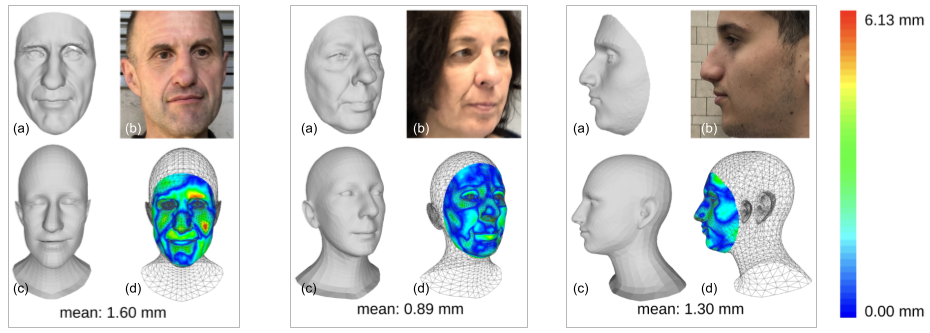}
  \vspace{1mm}
  \caption{\textbf{DAD-3DHeads accuracy} on selected samples from the NoW dataset. \textbf{(a)} GT scan; \textbf{(b)} input image; \textbf{(c)} the result of our annotation; \textbf{(d)} alignment of the mesh (wireframe) and the GT scan (with color-coded errors overlayed). 
  The scale of the errors relates to the real-world size of the scans. 
  Note that the resulting meshes accurately capture the coarse shape of the frontal part of the head,  
  the regions of higher error heavily overlap with finer facial structures. We provide more examples, in high resolution, in \cref{fig:now_scans_vs_GT_big} visualising this phenomenon. Best viewed zoomed in and in color.}
  \label{fig:now_GT_vis}
\end{figure*}

\textbf{3D Head Shape Reconstruction.} To validate that DAD-3DHeads annotations fit the head shape correctly, we compare the 3D meshes to the ground-truth scans provided in NoW \cite{RingNet} and Stirling \cite{Stirling} datasets, following the correspondent evaluation protocols (see \cref{ssec:3d_head_shape_reconstruction}).

As both benchmarks provide scans only of the frontal part of the face, the reconstruction of the whole skull can not be validated by those methods - that is where we resort to visual verification by our labelers as shown in \cref{fig:labeling_tool} (right).

We explicitly validate the accuracy on neutral images only since the 3D scans do not capture emotions, see the quantitative results in \cref{t:3dshape_now_GT}, \cref{t:3dshape_feng_GT}. For visual comparison, see \cref{fig:now_GT_vis}. Note that the representation is coarse (same as FLAME topology \cite{FLAME}), and we do not aim to model wrinkles and other tiny details that scanners can capture.
\begin{table}[t]
\footnotesize

\setlength{\tabcolsep}{3pt}
	\begin{subtable}[t]{.99\linewidth}\centering
		{

\centering
\tiny
\caption{\textbf{NoW\cite{RingNet} Dataset,} "multiview\_neutral" subset.}
\begin{adjustbox}{width=0.99\textwidth}
\begin{tabular}{@{}lccc@{}}

\hline %
    \textbf{Model} & \textbf{Median(mm)} & \textbf{Mean(mm)} & \textbf{Std(mm)}\\\hline %
    3DDFA-V2 \cite{3ddfa_cleardusk, guo2020towards} & 1.360 & 1.762 & 1.621                          \\
    RingNet \cite{RingNet} & 1.316 & 1.659 & 1.392\\
    DAD-3DHeads & \textbf{1.109} & \textbf{1.386} & \textbf{1.166}                          \\\hline
\noalign{\smallskip}
\end{tabular}
\end{adjustbox}

\label{t:3dshape_now_GT}}	
	\end{subtable}
	\hfill
	\begin{subtable}[t]{.99\linewidth}\centering
		{

\centering
\tiny
\caption{\textbf{Stirling\cite{Stirling} Database,} "Neutral expression, four views" subset.}
\begin{adjustbox}{width=0.99\textwidth}
\begin{tabular}{@{}lcccc@{}}

\hline %
    \textbf{Model} & \textbf{3DRMSE(mm)} & \textbf{Median(mm)} & \textbf{Mean(mm)} & \textbf{Std(mm)}\\\hline %
    RingNet \cite{RingNet} & 2.793 & 1.633 & 2.112 & 1.828 \\
    3DDFA-V2 \cite{3ddfa_cleardusk, guo2020towards} & 2.550 & 1.508 & 1.927 & 1.670         \\        DAD-3DHeads & \textbf{2.488} & \textbf{1.447} & \textbf{1.873} & \textbf{1.638}                          \\\hline
\noalign{\smallskip}
\end{tabular}
\end{adjustbox}

\label{t:3dshape_feng_GT}

}	
	\end{subtable}

\caption{\textbf{DAD-3DHeads accuracy of 3D Face Shape Reconstruction} on NoW and Stirling DBs; SOTA methods as reference.}
\end{table}


\textbf{3D Head Pose Estimation.} To validate the  goodness-of-fit of the head pose, we compare the rotation matrices from our annotations to the ground-truth matrices from the BIWI dataset \cite{BIWI}. They are captured by Kinect v2 sensors, the measurement error of which is 20mm \cite{Kinect_v2}.

To compare the matrices $R_1$ and $R_2$, we calculate the difference rotation $R_1 R_2^T$, and measure (i) Frobenius norm of the matrix $I - R_1 R_2^{T}$, as in \cite{rot_matrix_metrics}, and (ii) the angle in axis-angle representation of $R_1 R_2^T$, see \cref{t:3dpose_GT}.

\subsection{Annotation quality}\label{ssec:consistency}

To verify the quality of our annotations, we have selected a subset of $N=30$ images from different categories in the dataset. 
Each image was manually labeled with 68 facial landmarks, 
in the traditional configuration of \cite{multipie}, by $m=10$ different annotators. 
The same pictures were labeled following our annotation scheme (\cref{ssec:labeling}) in the 3D labeling tool. 
The 68 reprojected landmarks were computed from the 3D annotations to be comparable with the manual 2D-point labels. 
We compute the \textit{quality score} $\mathcal{F}_Q$ for each approach (see \Cref{t:consistency_matrix}), averaging across the images, as a normalized mean error between each pair of labels:
\begin{equation} \label{eq:422}
    \mathcal{F}_{Q} = \frac{1}{N} \sum_{n=1}^{N} \frac{1}{d_n} \cdot \frac{2}{m(m-1)} \sum_{i=1}^m \sum_{j>i} \Big|\Big|\overrightarrow{x_n^i} - \overrightarrow{x_n^j}\Big|\Big|_2,
\end{equation} 
where $d_n$ is the head bounding box size, as used in \cite{NME_bbox1, NME_bbox2}, $\overrightarrow{x}$ is an array of 68 labeled landmarks. 
As our data is mainly non-frontal, we do not use eye landmark distances as a normalization factor.
\definecolor{ao(english)}{rgb}{0.0, 0.5, 0.0}
\begin{table}[t]
\footnotesize
\renewcommand{\arraystretch}{0.95}
\centering
\begin{tabular}{r|c|c}
\noalign{\smallskip}
Method & $||I-R_1 R_2^{T}||_F$ & Angle error (degrees)\\
\hline
Img2Pose \cite{img2pose} & 0.228 & 9.336  \\
DAD-3DHeads & \textbf{0.149} & \textbf{6.037} \\ 
\noalign{\smallskip}
\end{tabular}
\caption{\textbf{DAD-3DHeads accuracy of 3D Head Pose estimation} on BIWI\cite{BIWI}; SOTA method as reference. The measure of $R_1 R_2^{T}$ deviation from identity matrix lies in the (0, $2\sqrt{2}$) range \cite{rot_matrix_metrics}.}
\label{t:3dpose_GT}
\end{table}

\begin{figure}[t]
\centering
\includegraphics[width=0.44\textwidth]{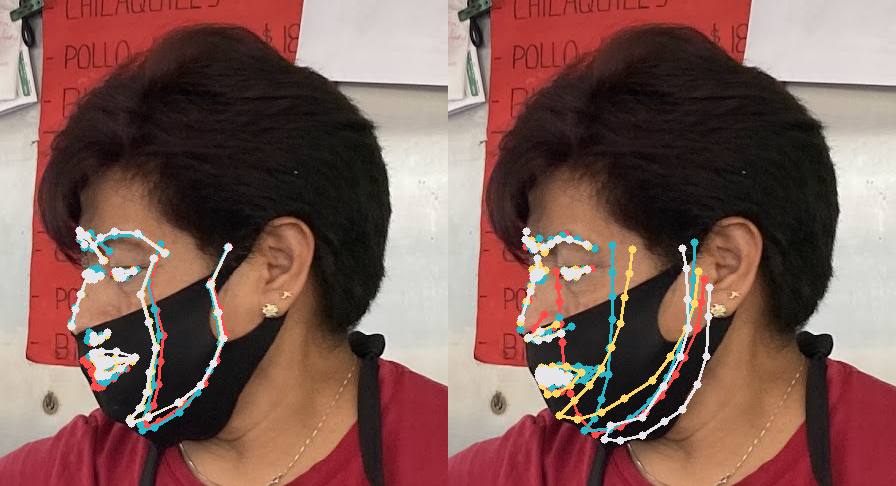}

  \caption{\textbf{Annotation Consistency.} Images labeled with our 3D annotation scheme (\textbf{left}) and with 68 2D points (\textbf{right}). Different colors correspond to different labelers. The annotators are consistent due to the conditioning by a 3D head model prior,
  which ensures high quality of the DAD-3DHeads dataset even under extremely diverse conditions. Labeling of invisible landmarks is highly inconsistent with the traditional approach, while using the 3D mesh fitting ensures high consistency even on occluded parts.
}
  \vspace{-1em}
  \label{fig:labeling_consistency}
\end{figure}

\definecolor{ao(english)}{rgb}{0.0, 0.5, 0.0}
\begin{table}[t]
\footnotesize
\renewcommand{\arraystretch}{0.95}
\centering
\begin{tabular}{r|c|c}
\noalign{\smallskip}
Method & $\mathcal{F}_Q$ (avg NME) & Best sample NME\\
\hline
2D 68 keypoints & 3.210 & 2.326\\
DAD-3DHeads 68 landm. & \textbf{1.737} \textcolor{ao(english)}{($\downarrow 45.8\%$)} &  \textbf{1.302} \\
\noalign{\smallskip}
\end{tabular}
\caption{\textbf{Quality score.} Annotation in 3D reduces  the global average NME by 45.8\%, see e.g. Fig. \ref{fig:labeling_consistency}.
}
\label{t:consistency_matrix}
\end{table}

\textbf{Limitations}. Such labelling scheme provides only partial control over depth. To mitigate this issue, we (i) provide the annotators with the ability to see the rendered texture onto the mesh in 3D, so they can inspect visually whether the lack of depth information corrupted the skull shape, and if the image provides realistic texture; (ii) propose $Z_n$ metric (see \cref{ssec:zn}) that assesses the depth quality.
\section{Method}

\begin{figure*}[t!]
\centering

  \includegraphics[width=\textwidth]{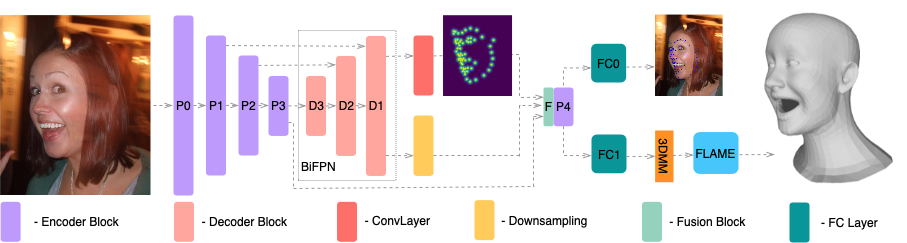}
  \vspace{-1em}
  \caption{\textbf{DAD-3DNet architecture design} and model training benefit from the rich annotations in a multi-branch setup. The Gaussian heatmap estimator predicts coarse locations of the head landmarks. The fusion block combines the coarse heatmap, BiFPN feature map, and CNN encoder output to regress a set of 3D head model parameters and finer locations of head landmarks.}
  \label{fig:arch}
\end{figure*}

Our goal is to estimate a compact 3D Head representation from a single image. Given an image, we assume the head is detected, loosely cropped, and approximately centered. We introduce a novel architecture, DAD-3DNet, that predicts a vector of 3DMM parameters disentangled into shape, expression and pose, and a dense set of 2D landmarks.
The landmarks serve as additional supervision and regularization and extend the range of applications that could benefit from the DAD-3DNet model. The DAD-3DNet architecture is illustrated in \cref{fig:arch}.

\subsection{DAD-3DNet Architecture}

Our architecture consists of (i) a CNN Encoder to extract features from the image, (ii) a Landmark Heatmap Estimator based on the BiFPN \cite{EfficientDet} to predict coarse locations of 2D landmarks, (iii) a Fusion Module that fuses the heatmap predictions with the encoder features, and (iv) a Regression Module that predicts finer facial landmarks locations and 3DMM parameters. 
We also use (v) a differential FLAME Layer that maps the 3DMM vector to the 3D mesh vertices. 

A pre-trained CNN Encoder extracts features from the first four stages of a backbone network. 
The Landmark Heatmap Estimator takes second to fourth stage feature maps as an input and predicts coarse Gaussian heatmaps using BiFPN, allowing easy and fast multi-scale feature fusion. 
The Gaussian heatmaps with $1/4$ of the original spatial resolution are then interpolated to the size of the fourth stage feature maps. 
The Fusion Layer incorporates the interpolated Gaussian heatmaps, the original feature map, and the BiFPN feature maps to encode a multi-scale representation with an Inception Module. 
A linear layer follows encoder representation to extract 2D landmarks locations. 

\subsection{Objective Function}\label{ssec:loss}

We introduce a multi-component loss function for the end-to-end training of DAD-3DNet to provide supervision for different branches of the network. The loss function consists of four different parts: \textbf{Shape+Expression Loss} measuring goodness-of-fit of the 3D Head Shape ($L_{3D}$), \textbf{Reprojection Loss} ($L_{proj}$) that incorporates pose information, Landmark Regression ($L_1$) and Gaussian Heatmap Loss ($L_{AWing}$ \cite{AWingLoss}) to provide the supervision for the 2D Facial Landmarks prediction branch. The detailed ablation studies (\cref{ssec:ablation}) show the importance of each component.

\textbf{Shape+Expression Loss:}\label{shapeLoss} Following the notations used in \cite{RingNet} , we denote the 3DMM coefficients as follows: shape coefficients $\overrightarrow{\beta}\in\mathbb{R}^{|\overrightarrow{\beta}|}$, expression coefficients $\overrightarrow{\psi}\in\mathbb{R}^{|\overrightarrow{\psi}|}$. 
The global rotation pose is modeled by $\overrightarrow{\theta_{r}}\in\mathbb{R}^{6}$ for continuity of representation \cite{continuity_rotation}, and is separated from the jaw rotation pose vector $\overrightarrow{\theta_{j}}\in\mathbb{R}^{3}$. 
In our approach we assume that neck $\overrightarrow{\theta_{n}}\in\mathbb{R}^{3}$ and eyeballs $\overrightarrow{\theta_{e}}\in\mathbb{R}^{6}$ rotation coefficients are equal to zero.
The global rotation predictions are set to zero to evaluate the discrepancy between our predictions and the ground truth in 3D. 
The 3D vertices are computed from the 3DMM parameters using a differentiable FLAME layer.
As FLAME model \cite{FLAME} contains both the head and the neck, but our task is narrowed down to the head mesh estimation, 
we subsample the vertices vector $\overrightarrow{v} = \overrightarrow{v}\big(\overrightarrow{\beta}, \overrightarrow{\psi}, \overrightarrow{\theta_{j}}\big)$ 
over the set of "head" vertex indices $I$: $\overrightarrow{v}|_I$.

The ground truth and the predicted mesh can differ in scale and location, so we normalize $\varphi$
both to fit into the unit cube after subsampling.

The final loss term measures discrepancy between normalized subsampled vertices:
\begin{equation} \label{eq424}
L_{3D}\Big(\overrightarrow{\beta}, \overrightarrow{\psi}, \overrightarrow{\theta_{j}}\Big) = \Big|\varphi\Big(\overrightarrow{v_{pred}|_I}\Big) - \varphi\Big(\overrightarrow{v_{GT}|_I}\Big)\Big|_2.
\end{equation}

\textbf{Reprojection Loss} is computed by projecting the 3D vertices of the posed mesh onto the image. 
The posed mesh is a "zero-pose" mesh described \hyperref[shapeLoss]{above}, to which we apply the similarity transform (rotation $R(\overrightarrow{\theta_{r}})$, uniform scaling $s$, and translation $\overrightarrow{t}$). 
The reprojection then is a simple orthographic projection onto the image plane. 
Here, as well, only the "head" vertices are included in the loss computation. 
The $L_1$ criterion is used as a discrepancy measure between the reprojected subsampled vertices. 
%

\textbf{The overall loss} is a combination of the four terms:\\
$\text{\quad\quad\quad}L = \lambda_1 L_{3D} + \lambda_2 L_1 + \lambda_3 L_{proj} + \lambda_4 L_{AWing}.$\\
We use 50.0, 1.0, 0.05, 1.0 as $\lambda_1$, $\lambda_2$, $\lambda_3$, $\lambda_4$ respectively. 






\subsection{Implementation details}
We implemented all of our models using PyTorch. The backbone network is initialized using the pre-trained weights on ImageNet. The differentiable FLAME layer is kept fixed during the training. The number of learnable head shape and expression parameters are set to 300 and 100, respectively. All the models are trained using \textbf{1 RTX A6000} GPU, with a batch size of 256. We use an ADAM optimizer with a learning rate = $1 * 10^{-4}$ and a plateau learning rate reducer with a reduce factor = 0.5 every six epochs when the validation loss stops decreasing. The training takes one day to converge. To preserve the scale ratio and shape of the head, images are padded to the square size and then resized to 256x256. We trained all models without any image augmentations. 

\section{Experimental Evaluation}

\begin{table*}[h!tb]
\footnotesize
\centering
\caption{\textbf{Comparison with state-of-the-art 3D Dense Head Alignment models on DAD-3DHeads Benchmark:} We compute the metrics on full test dataset as well as on challenging atypical poses (Pose), compound expressions (Expr.) and heavy occlusions (Occl.) subsets. DAD-3DNet shows superior performance on all subsets. Note: $Z_n$ is computed only for methods that use FLAME mesh topology.}.
\resizebox{\textwidth}{!} {
    \begin{tabular}{@{}l|cccc|cccc|cccc|cccc@{}}
    \toprule %
        \multirow{2}{*}{\textbf{Pose}}& 
        \multicolumn{4}{c}{\textbf{NME}{$\downarrow$}} &
        \multicolumn{4}{c}{\textbf{\textit{Z}\textsubscript{5}} \textbf{Accuracy}{$\uparrow$}} &
        \multicolumn{4}{c}{\textbf{Chamfer Distance}{$\downarrow$}} & 
        \multicolumn{4}{c}{\textbf{Pose Error}{$\downarrow$}} \\
                 & {Overall}  &  {Pose}  &  {Expr.}  &  {Occl.}    
                 & {Overall}  &  {Pose}  &  {Expr.}  &  {Occl.}  
                 & {Overall}  &  {Pose}  &  {Expr.}  &  {Occl.}  
                 & {Overall}  &  {Pose}  &  {Expr.}  &  {Occl.} \\\toprule %
        3DDFA-V2 \cite{3ddfa_cleardusk, guo2020towards}  & 3.580  &  7.630  &  3.168  &  3.195  &  
                   -  &  -  &  -  &  -  &  
                   6.17  &  8.878  &  6.410  &  6.400  & 
                   0.527  &  0.790  &  0.455  &  0.542   \\
        RingNet \cite{RingNet}  & 8.757  &  26.732  &  5.010  &  12.660  &  
                   0.880  &  0.743  &  0.913  &  0.860  &  
                   5.166  &  5.704  &  5.792  &  5.993  & 
                   0.438  &  1.076  &  	0.294  &  0.551   \\
        \textbf{DAD-3DNet}
                 & \textbf{2.302}  &  \textbf{6.049}  &  \textbf{1.748}  &  \textbf{2.036}  &  
                   \textbf{0.954}  &  \textbf{0.916}  &  \textbf{0.958}  &  \textbf{0.943}  &  
                   \textbf{3.178}  &  \textbf{4.094}  &  \textbf{3.375}  &  \textbf{3.774}  & 
                   \textbf{0.138}  &  \textbf{0.343}  &  \textbf{0.112}  &  \textbf{0.203}   \\\bottomrule
    \end{tabular}}
\label{T:dad_benchmark}
\end{table*}

We propose DAD-3DHeads Benchmark for evaluating (i) the task of 3D Dense Head Alignment from an image, (ii) in-the-wild model generalization (when trained on our data) to a range of 3D Head Learning tasks, and (iii) robustness to extreme poses. 
To address (i), we provide a comprehensive analysis of DAD-3DNet and several existing methods on our benchmark, and report the findings in \cref{T:dad_benchmark}. 
To test generalization (ii), we analyze the performance of DAD-3DNet on the established benchmarks for 3D Face Shape Reconstruction and 3D Head Pose Estimation, detailed in \cref{ssec:3d_head_pose_estimation}, \cref{ssec:3d_head_shape_reconstruction}.
To test robustness (iii), we evaluate DAD-3DNet under train/test distribution shift in camera poses and report our findings in \cref{ssec:distribution_shifts} (see \cref{T:allmetrics}).
\subsection{Metrics}\label{ssec:benchmark}
Given a ground-truth mesh $M$ on a particular frame, and post-processed model output - predicted 3D vertices $V$, we calculate how well $V$ fit $M$. 
The goodness-of-fit measures the pose fitting, and both face and head shape matching. 
We propose two new metrics for the evaluation protocol: Reprojection NME and $Z_{n}$ accuracy, in addition to Chamfer Distance and Pose Error reported previously for the 3D Head Learning tasks.

\textbf{Reprojection NME:} we compute the normalized mean error of the reprojected 3D vertices onto the image plane, taking $X$ and $Y$ coordinates into account. Similar to \cref{eq:422}, we use head bounding box size for normalization. The metric is computed on 68 landmarks \cite{multipie}.

\textbf{\textit{Z\textsubscript{n}} Accuracy:}\label{ssec:zn} as our annotation scheme is conditioned only upon model prior and the reprojection onto the image, we do not guarantee the absolute depth values to be as accurate as sensor data.
We address this issue by measuring the \emph{relative depth} as an ordinal value of the $Z$-coordinate. 
For each of $K$ vertices $v_i$ of the GT mesh, we choose $n$ closest vertices $\{v_i^1, ..., v_i^n\}$ , and calculate which of them are closer to (or further from) the camera. Then, we compare if for every predicted vertex $w_i$ this configuration is the same:\\
$\text{\quad}Z_n = \frac{1}{K}\frac{1}{n}\sum_{i=1}^K \sum_{j=1}^n\Big((v_i \succeq_z v_i^j) == (w_i \succeq_z w_i^j)\Big).$
We do so on the "head" subset of the vertices only.

\textbf{Chamfer Distance:} as the $Z_n$ metric is valid only for predictions that follow FLAME mesh topology, we add Chamfer distance to measure the accuracy of fit. To ensure generalization to any number of predicted vertices, we measure a one-sided Chamfer distance from our ground-truth mesh to the predicted one. We align them by seven key points correspondences \cite{RingNet}, and compute the distances from the "face" subset of the vertices only (see \cref{fig:head_and_face} in Appendix), following the traditional approach \cite{RingNet, 3d-face-modeling-from-diverse-raw-scan-data}.

\begin{figure}[t]
\centering
\includegraphics[width=0.47\textwidth]{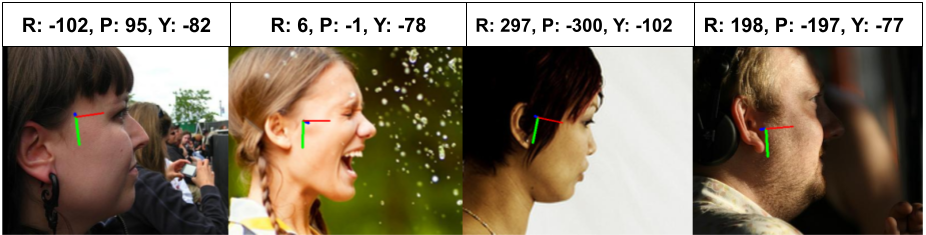}

  \caption{\textbf{AFLW2000-3D label inconsistencies.} Some labels of side or extreme atypical poses are inconsistent as the Euler angle representation used is ambiguous due to gimbal lock.}
  \vspace{-1em}
  \label{fig:paradox}
\end{figure}

\textbf{Pose Error:} measuring the accuracy of pose prediction, we want to overcome the issues observed in AFLW2000-3D\cite{koestinger11a} Dataset. Creators of AFLW2000-3D measure the head pose resorting to Euler angles. Such representation is highly dependent on the order in which the rotations are applied. Whenever the second rotation reaches over $\frac{\pi}{2}$ in any direction, i.e., extreme head poses, other rotation axes become linearly dependent, yielding an infinite number of representations for the same transformation \cite{eruler_rotation}.
One can observe inconsistencies caused by this in the AFLW2000-3D\cite{koestinger11a} benchmark in \cref{fig:paradox}.

To avoid that, we measure accuracy of pose prediction based on rotation matrices \cite{rot_matrix_metrics} (see \cref{ssec:accuracy_labels}):\\
$\text{\quad\quad\quad\quad}Error_{pose} = ||I-R_1 R_2^{T}||_F$

\subsection{3D Head Pose Estimation}\label{ssec:3d_head_pose_estimation}

We evaluate DAD-3DNet on AFLW2000-3D and BIWI datasets for the task of 3D Head Pose Estimation.

\definecolor{Gray}{gray}{0.9}
\textbf{BIWI Dataset} \cite{BIWI} is gathered in a laboratory setting by recording RGB-D video of different subjects across various head poses using a Kinect v2 device. 
It contains frames with the rotations ±75$^{\circ}$ for yaw, ±60$^{\circ}$ for pitch, and ±50$^{\circ}$ for roll. 
A 3D model was fit to each individual's point cloud, and the head rotations were tracked to produce the pose annotations. 

\textbf{AFLW2000-3D Dataset}  \cite{AFLW2000-3D} consists of the first 2,000 subjects of the in-the-wild AFLW dataset, which has been re-annotated with image-level 68 3D landmarks
and consequently, contain fine-grained pose annotations. 

\textbf{Results:} We report the results in \cref{t:3dpose_biwi}, \cref{t:3dpose_aflw}.
The proposed model outperforms all other 3DMM estimation methods by a large margin, and shows comparable performance to other state-of-the-art methods for head pose estimation. 

\begin{table}[t]
\footnotesize
\caption{\textbf{3D head pose estimation} results.}
\setlength{\tabcolsep}{3pt}
	\begin{subtable}[t]{.99\linewidth}\centering
		{\scriptsize
\begin{adjustbox}{width=0.99\textwidth}
\begin{tabular}{@{}lrrrr@{}}
\noalign{\smallskip}
\toprule %
\textbf{Model} & \textbf{MAE} ↓ & \textbf{Pitch MAE} ↓ & \textbf{Roll MAE} ↓ & \textbf{Yaw MAE} ↓ \\\toprule %
3DDFA \cite{3ddfa_cleardusk, zhu2017face} \ & 19.07 & 12.25 & 8.78 & 36.18\\
Fan (12 points)\cite{FacePoseNet} & 7.88 & 7.48 & 7.63 & 8.53\\
Dlib (68 points)\cite{dlib09} & 12.25 & 13.80 & 6.19 & 16.76\\
HopeNet\cite{HOPE-Net} & 4.90 & 6.61 & 3.27 & 4.81\\
Img2Pose\cite{img2pose} & \textbf{3.79} & \textbf{3.55} & 3.24 & 4.57\\
3DDFA-V2\cite{3ddfa_cleardusk, guo2020towards} & 8.81 & 12.08 & 7.54 & 6.80 \\
RingNet\cite{RingNet} & 7.34 & 5.37 & 7.82 & 8.82 \\
WHENet\cite{WHENet} & 3.81 & 4.39 & 3.06 & 3.99 \\
\textbf{DAD-3DNet} & 3.98 & 5.24 & \textbf{2.92} & \textbf{3.79}\\\bottomrule %
\noalign{\smallskip}
\end{tabular}
\end{adjustbox}
\caption{\textbf{BIWI}\cite{BIWI}}
\label{t:3dpose_biwi}}	
	\end{subtable}
	\hfill
	\begin{subtable}[t]{.99\linewidth}\centering
		{\begin{adjustbox}{width=0.99\textwidth}
\begin{tabular}{@{}lrrrr@{}}
\noalign{\smallskip}
\toprule %
\textbf{Model} & \textbf{MAE} ↓ & \textbf{Pitch MAE} ↓ & \textbf{Roll MAE} ↓ & \textbf{Yaw MAE} ↓ \\\toprule %
3DDFA\cite{3ddfa_cleardusk, zhu2017face} & 7.39 & 8.53 & 7.39 & 5.40\\
Fan (12 points)\cite{FacePoseNet} & 9.12 & 12.28 & 8.71 & 6.36\\
Dlib (68 points)\cite{dlib09} & 13.29 & 12.60 & 9.00 & 18.27\\
HopeNet\cite{HOPE-Net} & 6.16 & 6.56 & 5.44 & 6.47\\
RetinaNet\cite{RetinaFace} & 6.22 & 9.64 & 3.92 & 5.10\\
Img2Pose\cite{img2pose} & 3.91 & 5.03 & 3.28 & 3.43\\
SynergyNet\cite{wu2021synergy} & \textbf{3.35} & \textbf{4.09} & \textbf{2.55} & 3.42\\
3DDFA-V2\cite{3ddfa_cleardusk, guo2020towards} & 7.56 & 8.48 & 9.89 & 4.30\\
RingNet\cite{RingNet} & 8.27 & 4.39 & 13.51 & 6.92\\
\textbf{DAD-3DNet} & 3.66 & 4.76 & 3.15 & \textbf{3.08}\\\bottomrule %
\noalign{\smallskip}
\end{tabular}
\end{adjustbox}
\caption{\textbf{AFLW2000-3D}\cite{AFLW2000-3D}}
\label{t:3dpose_aflw}}	
	\end{subtable}
\vspace{-2.0em}
\label{t:3dpose}
\end{table}
\hspace{2em}

\subsection{3D Face Shape Reconstruction}\label{ssec:3d_head_shape_reconstruction}
For the task of 3D Face Shape reconstruction, we compare the performance of DAD-3DNet with two state-of-the-art publicly available methods: 3DDFA-V2 \cite{3ddfa_cleardusk, guo2020towards},  and RingNet \cite{RingNet} on two 3D Face Shape reconstruction benchmarks: NoW \cite{RingNet} and Feng et al. \cite{3d-face-modeling-from-diverse-raw-scan-data}.

\textbf{NoW Face Challenge:} 
NoW benchmark is designed for the task of 3D face reconstruction from single monocular images. The dataset contains 2054 2D images of 100 subjects. Following the evaluation protocol, we predict the meshes that are then rigidly aligned with corresponding ground truth scans based on seven landmark points. The scan-to-mesh distance is computed between them. The calculated mean, median, and standard deviation errors are reported in \Cref{t:3dshape_now}.
\begin{table}[t]
\bigskip
\footnotesize
\caption{\textbf{3D face shape reconstruction} results. }
\setlength{\tabcolsep}{3pt}
	\begin{subtable}[t]{.99\linewidth}\centering
		{
\centering
\tiny
\begin{adjustbox}{width=0.99\textwidth}
\begin{tabular}{@{}lccc@{}}

\hline %
    \textbf{Model} & \textbf{Median(mm)} ↓ & \textbf{Mean(mm)} ↓ & \textbf{Std(mm)} ↓ \\\hline %
    3DDFA-V2\cite{3ddfa_cleardusk, guo2020towards} & 1.234 & 1.566 & 1.391                          \\
    RingNet\cite{RingNet} & \textbf{1.207} & \textbf{1.535} & 1.306\\
    \textbf{DAD-3DNet} & 1.236 & 1.541 & \textbf{1.285}                        \\\hline
\noalign{\smallskip}
\end{tabular}
\end{adjustbox}
\caption{\textbf{NoW\cite{RingNet}}}
\label{t:3dshape_now}}	
	\end{subtable}
	\hfill
	\begin{subtable}[t]{.99\linewidth}\centering
		{\footnotesize
\begin{adjustbox}{width=0.99\textwidth}
\begin{tabular}{@{}lccccccc@{}}
\noalign{\smallskip}
\toprule %
    \multirow{2}{*}{\textbf{Model}}& \multirow{2}{*}{\textbf{3DRMSE}↓} & 
    \multicolumn{2}{c}{\textbf{Median(mm)}↓} &
    \multicolumn{2}{c}{\textbf{Mean(mm)}↓} & 
    \multicolumn{2}{c}{\textbf{Std(mm)}↓} \\
             &       &   HQ  &   LQ  &   HQ  &   LQ  &   HQ  &   LQ  \\\toprule %
    3DDFA-V2\cite{3ddfa_cleardusk, guo2020towards} &   2.998 &   \textbf{1.500}  &   1.779  &   1.942   &  2.350    &   1.704  & 2.149  \\
    RingNet\cite{RingNet}  &   2.809   &   1.698   &   1.634   &   2.161  &   2.113    &  1.832   & 1.831  \\
    \textbf{DAD-3DNet} &   \textbf{2.749}   &   1.558   &   \textbf{1.624}   &   \textbf{1.940}   &   \textbf{2.082}   &   \textbf{1.581}   &   \textbf{1.795}   \\\bottomrule
\noalign{\smallskip}
\end{tabular}
\end{adjustbox}
\caption{\textbf{Feng et al.\cite{3d-face-modeling-from-diverse-raw-scan-data}}}
\label{t:3dshape_feng}}	
	\end{subtable}
\vspace{-2.0em}
\end{table}

\textbf{Feng et al. Benchmark:} \cite{3d-face-modeling-from-diverse-raw-scan-data} provides a subset of Stirling/ESRC 3D face database as the test dataset for their challenge. The test dataset consists of 2,000 2D various expression facial images, including 656 high-quality (HQ) images taken in controlled scenarios and 1,344 low-quality (LQ) images extracted from video frames \cite{Stirling}. Following \cite{3d-face-modeling-from-diverse-raw-scan-data} protocol that rehearses \cite{RingNet} we perform the aforementioned steps and  compute the scan-to-mesh distance between the predicted meshes and the ground truth scans. These distances are used to compute the 3DRMSE. We also compute mean, median and standard deviation errors for HQ and LQ images separately for in-depth analysis. The results of evaluation are provided in the \Cref{t:3dshape_feng}.

\textbf{Results:} DAD-3DNet shows superior performance to other methods for coarse 3D dense head alignment without using explicit Shape and Expression disentanglement loss. 


\subsection{Ablation study}
\label{ssec:ablation}

In this section, we verify the efficiency of the separate loss components and demonstrate the impact of the training data. We report the results of an ablation study in Table \ref{t:ablation}. 

\begin{table}[h!]
\footnotesize
\renewcommand{\arraystretch}{0.95}
\centering
\begin{tabular}{r|l|ccc}
\noalign{\smallskip}
 & \textbf{Component} & \textbf{NME}{$\downarrow$} & {\textbf{\textit{Z}\textsubscript{5}} \textbf{Acc.}{$\uparrow$}} & \textbf{Pose}{$\downarrow$}\\
\hline
1 & \textit{baseline} & 2.576 & 0.880 & 0.267\\
2 & + full face reprojection loss & 2.395 & 0.873 & 0.263\\
3 & + full head reprojection loss & 2.500 & 0.943 & 0.172\\
4 & + shape+expression loss & 2.471 & 0.951 & 0.139 \\
5 & + landmark prediction head & \textbf{2.302} & \textbf{0.954} & \textbf{0.138} \\
\noalign{\smallskip}
\end{tabular}
\caption{ \textbf{DAD-3DNet ablation study on DAD-3DHeads:} 
The loss terms have significant impact on the fitting accuracy, and the multi-head architecture improves the model generalization.}
\label{t:ablation}
\end{table}

\textbf{Reprojection Loss:} Supervision based on reprojected landmarks is a core part of the training algorithms. Compared to the models that use supervision based on 68 keypoints, we have only added the reprojection loss based on all available face and head points. Incorporating information about other facial landmarks improves the accuracy of reprojected 68 landmarks but does not impact the other metrics and does not improve the 3D fitting; adding points of  the whole head improved all the metrics by a large margin. Additional full head supervision improves the model stability by enforcing to learn the entire head shape.

\textbf{Shape+Expression Loss:} Rich supervision of the normalized 3D vertices locations enables the model to encode more nuanced information about the 3D head pose. As shown in Table \ref{t:ablation} this component improves all of the metrics and reduces the 3D head pose error significantly.

\textbf{Landmarks Head:} Multi-task training improves the model stability and enforces the model to prefer more general representations. With the landmarks regression and coarse heatmap estimation modules, the model achieves a significant boost in performance on all metrics yet again.

\section{Conclusions}
We introduce DAD-3DHeads, a dense, accurate, and diverse 3D Head dataset in the wild. We demonstrate the efficiency and accuracy of the data and novel loss components by training a data-driven DAD-3DNet model. DAD-3DNet achieves superior performance on diverse 3D head tasks and successfully generalizes to in-the-wild conditions. 

\textbf{Acknowledgements.} We thank the Armed Forces of Ukraine for providing security to complete this work.

{\small
\bibliographystyle{ieee_fullname}
\bibliography{egbib}
}
\clearpage

\appendix



\section{DAD-3DHeads Dataset}

The images in DAD-3DHeads dataset are anonymised without additional metadata. 
The results of labeling in the form of 3D head model do not contain any private or sensitive information.
The data gathered is not being used for identification purposes or in connection with any other personal data.

\label{sec:dataset}
\subsection{Dataset card}\label{ssec:dataset_card_appendix}
\begin{figure*}[t!]
     \centering
      \includegraphics[width=0.9\textwidth]{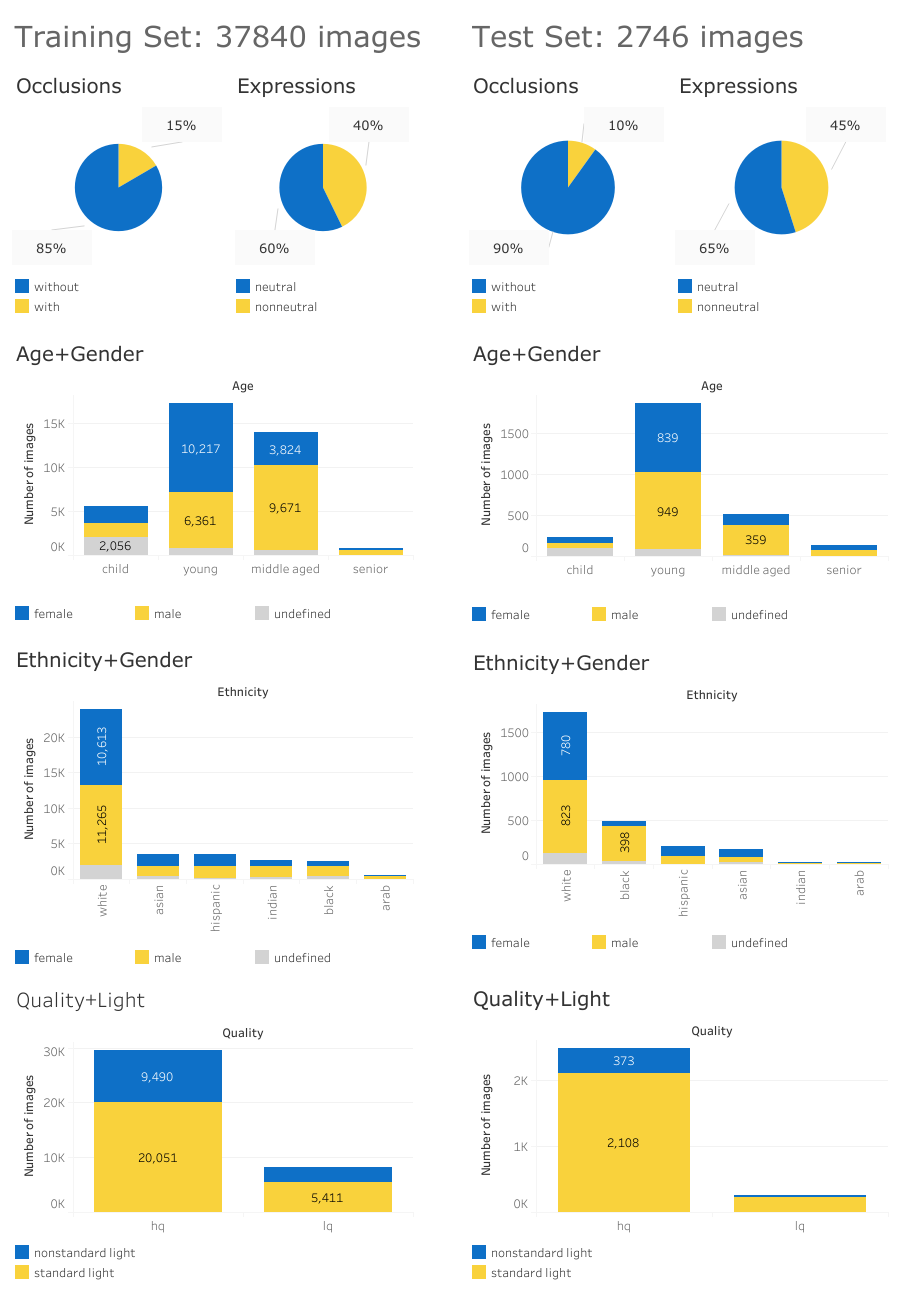}
     \caption{Attribute labels (gender, age, illumination, and image quality) and the distribution across ethnic groups in DAD-3DHeads.}
     \label{fig:histogramss}
\end{figure*}

As stated in \cref{ssec:data_statistics}, along with the dataset of images and annotations, we provide additional information per image such as gender, age, illumination conditions, image quality, pose, presense of expression, and occlusions (see \cref{fig:histogramss} and \cref{fig:dataset_stats}). 


We use multiple sources to construct our dataset, among which WIDER FACE dataset \cite{wider},
the Adience dataset \cite{adience}, Compound Facial Expressions of Emotions Dataset\cite{fce}, 
WFLW\cite{wflw}, AFW\cite{afw}, Helen\cite{helen}, LFPW\cite{LFPW}.



 


\subsection{DAD-3DHeads annotations accuracy}

\begin{figure*}[t!]
  \centering
  \includegraphics[width=0.85\textwidth]{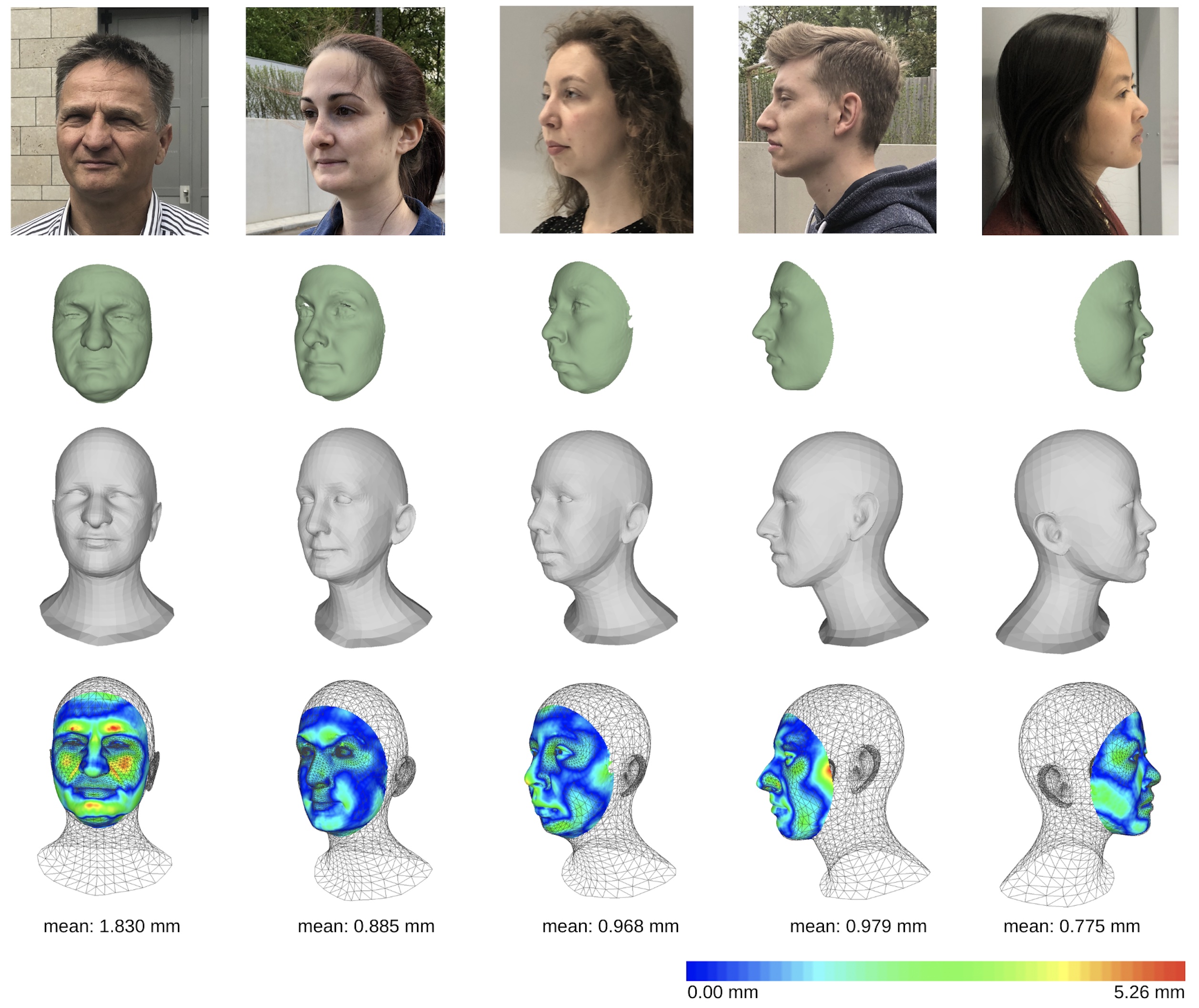}
  \vspace{1mm}
  \caption{\textbf{DAD-3DHeads accuracy} on selected samples from the NoW dataset. \textbf{First row:} input image; \textbf{second row:} GT scan; \textbf{third row:} the result of our annotation; \textbf{fourth row:} alignment of the mesh (wireframe) and the GT scan (with color-coded errors overlayed). 
}
  \label{fig:now_scans_vs_GT_big}
\end{figure*}

We provide more visualizations of the results of DAD-3DHeads annotations compared to GT scans on neutral faces from NoW dataset\cite{RingNet} in \cref{fig:now_scans_vs_GT_big}. This is in addition to (and in the context of) \cref{fig:now_GT_vis}.

\subsection{Annotation process}\label{ssec:annotation_process_detailed}

We provide here more details along with visuals to prevent possible misunderstandings around the labeling process. The full video is available via \href{https://p.farm/research/dad-3dheads}{the project webpage}.

\begin{figure}[h!]
     \centering
      \includegraphics[width=0.45\textwidth]{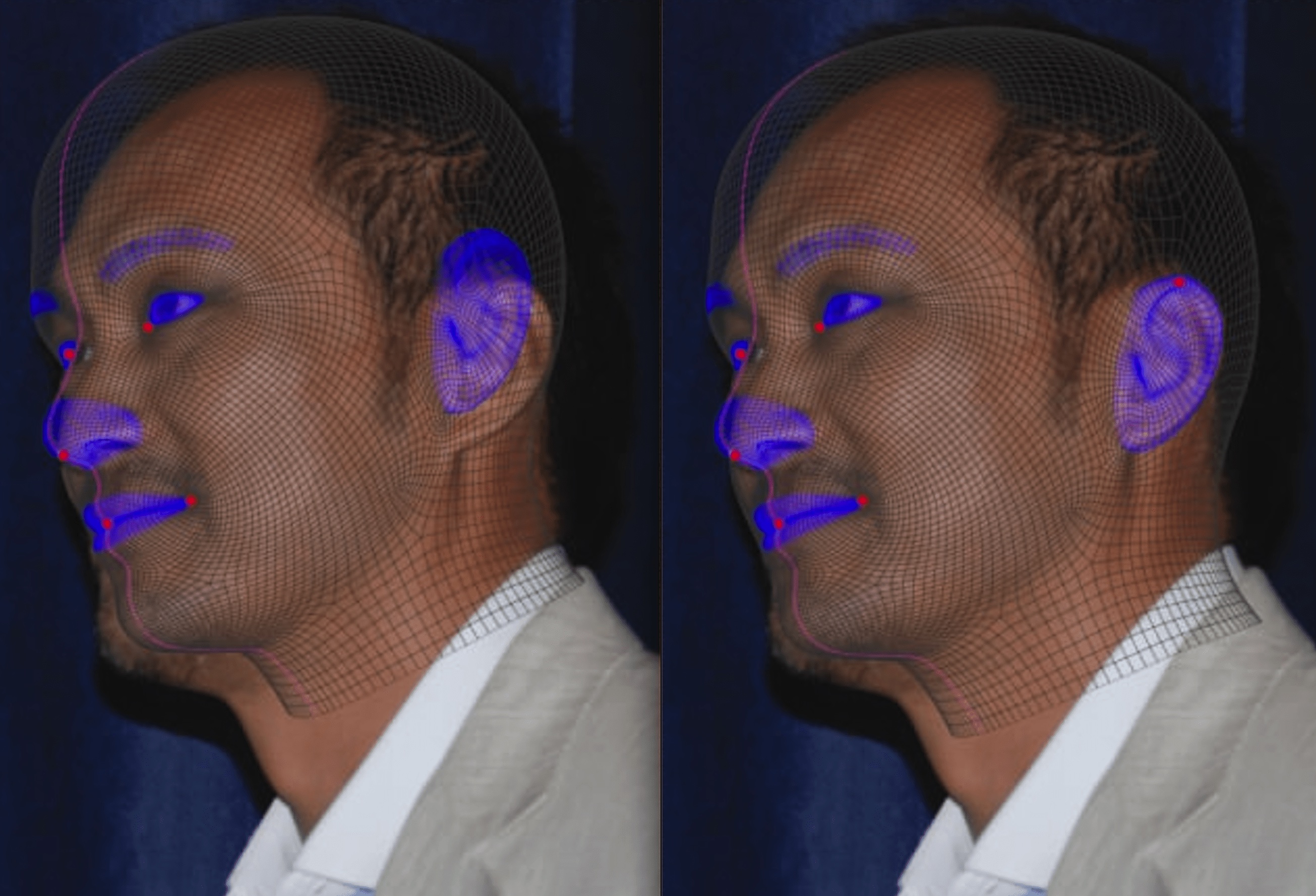}
     \caption{The mesh is being deformed due to the under-the-hood optimization after the "pin" is placed on the ear.}
     \label{fig:iteration}
\end{figure}

As was already mentioned in the \cref{ssec:labeling}, the annotators do not explicitly control or label either the 3DMM parameters or the blendshapes. 
They also do not have to disambiguate between identity and expression.
The annotators only have to "pin" a 3D head model - a mesh - to the head image.
They do so iteratively having the initial generic mesh optimized after every "pin" being placed (see \cref{fig:iteration}).
The nonlinear optimization is performed, indeed, over the shape and expression parameters, along with the pose.
\begin{figure}[h!]
     \centering
      \includegraphics[width=0.45\textwidth]{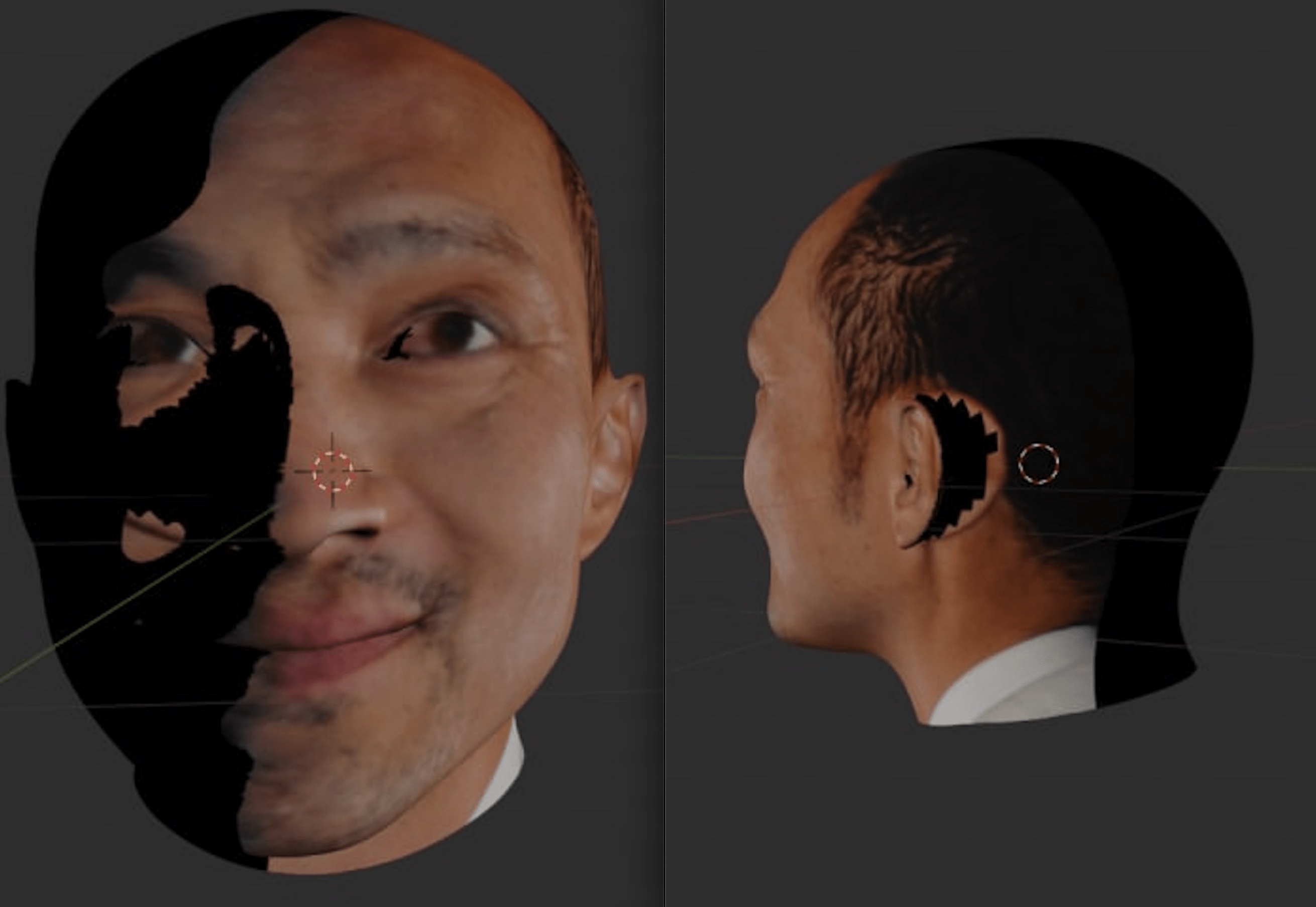}
     \caption{The rendered texture here has "holes" due to the occlusions, but can also be "torn" if the "pins" are placed poorly.}
     \label{fig:holes}
\end{figure}

After each step, the annotators can inspect if the fitted mesh aligns well with the image in several ways: they might view (i) the reprojected set of selected landmarks that correspond to the recognizible features on any face or head (see \cref{fig:191_landm} as an example), such as contours of the eyes (eyelash line), lips, etc.; (ii) the image rendered onto the mesh as a texture given the fitting - it helps to inspect if there appeared texture "holes" due to poor labeling or occlusions (see \cref{fig:holes}); (iii) the mesh itself is visible in $360^o$ to inspect if the skull shape is not deformed (see \cref{fig:holes}, right).
These measures help to partially overcome limitations introduced by the absence of camera parameters for the input images, thus possible ambiguities caused by the effects of perspective projection.

Another important issue is that our annotations consist only of the 3D vertices and transformation matrices, we do not interpret the resulting mesh w.r.t. identity, expression, or any other ambiguous and ill-defined over a single-image input feature.
The concerns arise as there might be cases, e.g., faces with extreme expressions, where it is impossible to perfectly detect the shape of the head with neutral expression based on a single image.
Indeed, the 3D scanners would provide an accurate 3D head model that the manual annotation cannot guarantee. 
However, they operate under controlled capture, i.e., \emph{3D scans have been in-the-lab instead of in-the-wild}. 
We provide the community with the complementary data. 
It has a \emph{known} trade-off in accuracy, but the performance is sufficient for many applications that operate in-the-wild.

\section{Experimental results on DAD-3DHeads benchmark}

\subsection{3D Landmark localization}

We evaluate the state-of-the-art methods such as JVCR\cite{jvcr}, FaceSynthetics\cite{wood2021fake}, 3DDFA-v2\cite{guo2020towards}, and proposed DAD-3DNet on DAD-3DHeads benchmark for the task of 3D Landmark Localization. We report normalized mean error (NME) for the predicted landmarks reprojected onto the image plane (see \cref{t:3d_landarks_dad}). 
We analyse the NME metric on full test dataset as well as across challenging subgroups (atypical poses, compound expressions, heavy occlusions). \emph{DAD-3DNet shows superior performance in all cases.}

When computing NME of the competitor methods, we only use the images where the landmarks have been localized, otherwise the NME is ill-defined. See examples where the landmarks are not localized in \cref{fig:3d_landmarks_dad_a}, along with other challenging cases.

Performance on the "Expr." subset is better than the overall average (\Cref{t:3d_landarks_dad}) for all of the methods. We attribute this to the fact that for heavily occluded faces and large extreme poses, where the landmarks are not clearly visible (and therefore the emotions), the images are labeled as "neutral" by default.

\begin{table}[t]
\footnotesize
\centering

    \begin{tabular}{@{}l|cccc@{}}
    \toprule %
        \multirow{2}{*}{\textbf{Model}}& 
        \multicolumn{4}{c}{\textbf{NME}{$\downarrow$}}\\
                 & {Overall}  &  {Pose}  &  {Expr.}  &  {Occl.} \\\toprule %
        3DDFA-V2 \cite{guo2020towards}  & 3.580  &  7.630  &  3.168  &  3.195  \\
        FaceSynthetics \cite{wood2021fake}  & 4.363  &  15.781  &  3.159  &  	4.934  \\
        JVCR \cite{jvcr}  & 4.455  &  12.514  &  3.843  &  4.949  \\

        \textbf{DAD-3DNet}
                 & \textbf{2.302}  &  \textbf{6.049}  &  \textbf{1.748}  &  \textbf{2.036}  \\\bottomrule
    \end{tabular}
\caption{\textbf{3D Landmark Localization on DAD-3DHeads benchmark.} We compute the normalized mean error (NME, the lower the better) on full test dataset as well as on challenging atypical poses (Pose), compound expressions (Expr.) and heavy occlusions (Occl.) subsets. DAD-3DNet performs superior in all cases.}.
\label{t:3d_landarks_dad}
\end{table}


\subsection{DAD-3DNet performance across multiple subgroups}\label{ssec:distribution_shifts}
We present an extension of \cref{T:dad_benchmark} from the main paper, performing in-depth analysis of the results across multiple subgroups. We define subgroups w.r.t. \emph{pose} (front, side, atypical), \emph{age} (child, young, middle age, senior), \emph{image quality} (high, low), \emph{occlusions} (true, false), \emph{expressions} (neutral, non-neutral), \emph{lightning} (standard, non-standard). The results of the DAD-3DNet model are reported in \cref{T:allmetrics}. This analysis shows robustness of the proposed approach across various conditions (distribution shifts) in-the-wild.

\begin{table}
\footnotesize
\centering
\label{T:allmetrics}
\begin{tabular}{l|cccc} 
\toprule
\multicolumn{1}{c}{\textbf{ Pose}} & \textbf{\textbf{NME}} & \textbf{\textbf{\textit{Z}5}}~\textbf{\textbf{Accuracy}} & \textbf{\textbf{Chamfer Dist.}} & \multicolumn{1}{l}{\textbf{\textbf{Pose Error}}}  \\ 
\toprule
front                              & 1.496                 & 0.965                                                    & 3.146                              & 0.089                                             \\
side                              & 2.257                 & 0.952                                                    & 3.180                              & 0.143                                             \\
atypical                           & 6.190                 & 0.916                                                    & 4.027                              & 0.343                                             \\
\bottomrule
\end{tabular}
\footnotesize
\label{T:allmetrics}
\begin{tabular}{l|cccc} 
\toprule
\multicolumn{1}{c}{\textbf{Age}} & \textbf{\textbf{NME}} & \textbf{\textbf{\textit{Z}5}}~\textbf{\textbf{Accuracy}} & \textbf{\textbf{Chamfer Dist.}} & \multicolumn{1}{l}{\textbf{\textbf{Pose Error}}}  \\ 
\toprule
child                            & 1.662                 & 0.960                                                    & 3.546                          & 0.103                                             \\
young                            & 2.421                 & 0.953                                                    & 3.178                          & 0.150                                             \\
middle                     & 2.438                 & 0.955                                                    & 3.393                          & 0.133                                             \\

senior                           & 1.756                 & 0.958                                                    & 2.989                          & 0.113                                             \\
\bottomrule
\end{tabular}
\footnotesize
\label{T:allmetrics}
\begin{tabular}{l|cccc} 
\toprule
\multicolumn{1}{c}{\textbf{Quality}} & \textbf{\textbf{NME}} & \textbf{\textbf{\textit{Z}5}}~\textbf{\textbf{Accuracy}} & \textbf{\textbf{Chamfer Dist.}} & \multicolumn{1}{l}{\textbf{\textbf{Pose Error}}}  \\ 
\toprule
high                                 & 2.065                 & 0.957                                                    & 3.194                          & 0.129                                             \\
low                                  & 4.755                 & 0.928                                                    & 3.643                          & 0.259                                             \\
\bottomrule
\end{tabular}
\footnotesize
\label{T:allmetrics}
\begin{tabular}{l|cccc} 
\toprule
\multicolumn{1}{c}{\textbf{Occlusion}} & \textbf{\textbf{NME}} & \textbf{\textbf{\textit{Z}5}}~\textbf{\textbf{Accuracy}} & \textbf{\textbf{Chamfer Dist.}} & \multicolumn{1}{l}{\textbf{\textbf{Pose Error}}}  \\ 
\toprule
True                                   & 4.242                 & 0.9436                                                   & 3.784                           & 0.1986                                            \\
False                                  & 2.134                 & 0.955                                                    & 3.182                           & 0.1359                                            \\
\bottomrule
\end{tabular}
\footnotesize
\label{T:allmetrics}
\begin{tabular}{l|cccc} 
\toprule
\multicolumn{1}{c}{\textbf{Expression}} & \textbf{\textbf{NME}} & \textbf{\textbf{\textit{Z}5}}~\textbf{\textbf{Accuracy}} & \textbf{\textbf{Chamfer Dist.}} & \multicolumn{1}{l}{\textbf{\textbf{Pose Error}}}  \\ 
\toprule
Non-neutral                                    & 1.156                 & 0.959                                                    & 3.412                           & 0.116                                             \\
Neutral                                   & 1.644                 & 0.950                                                    & 3.088                           & 0.164                                             \\
\bottomrule
\end{tabular}
\footnotesize
\begin{tabular}{l|cccc} 
\toprule
\multicolumn{1}{c}{\textbf{Lighting}} & \textbf{\textbf{NME}} & \textbf{\textbf{\textit{Z}5}}~\textbf{\textbf{Accuracy}} & \textbf{\textbf{Chamfer Dist.}} & \multicolumn{1}{l}{\textbf{\textbf{Pose Error}}}  \\ 
\toprule
Standard                                        & 2.350                 & 0.954                                                    & 3.182                           & 0.142                                             \\
Non-standard                                       & 2.235                 & 0.954                                                    & 3.569                           & 0.142                                             \\
\bottomrule
\end{tabular}
\caption{In-depth analysis of the benchmark results reported in Table 4 of the main paper across multiple subgroups such as \emph{camera pose}, \emph{age}, \emph{image quality}, \emph{occlusions}, \emph{expressions}, \emph{lighting}. This analysis shows robustness of the proposed approach across various conditions (distribution shifts) in-the-wild.}
\label{T:allmetrics}
\end{table}

\subsection{3D Head Pose Estimation}

The comparison of DAD-3DNet with the state-of-the-art 3D Head Pose Estimation method img2pose \cite{img2pose} is provided in \Cref{t:3dpose_img2pose_we}. We calculate two metrics on the rotation matrices $R_1$ (GT) and $R_2$ (prediction), similar to \Cref{t:3dpose_GT}: (i) Frobenius norm of the matrix $I - R_1 R_2^{T}$, and (ii) the angle in axis-angle representation of $R_1 R_2^T$.


\definecolor{ao(english)}{rgb}{0.0, 0.5, 0.0}
\begin{table}[t]
\footnotesize
\renewcommand{\arraystretch}{0.95}
\centering
\begin{tabular}{r|c|c}
\noalign{\smallskip}
Method & $||I-R_1 R_2^{T}||_F$ & Angle error (degrees)\\
\hline
Img2Pose \cite{img2pose} & 0.226  & 9.122  \\
DAD-3DNet & \textbf{0.138} & \textbf{5.360} \\ 
\noalign{\smallskip}
\end{tabular}
\caption{\textbf{3D Head Pose estimation on DAD-3DHeads benchmark}. DAD-3DNet outperforms state-of-the-art img2pose\cite{img2pose}. The measure of $R_1 R_2^{T}$ deviation from identity matrix lies in the (0, $2\sqrt{2}$) range \cite{rot_matrix_metrics}.}
\label{t:3dpose_img2pose_we}
\end{table}

\subsection{Failure cases}
\begin{figure}[h]
\centering
\includegraphics[width=0.45\textwidth]{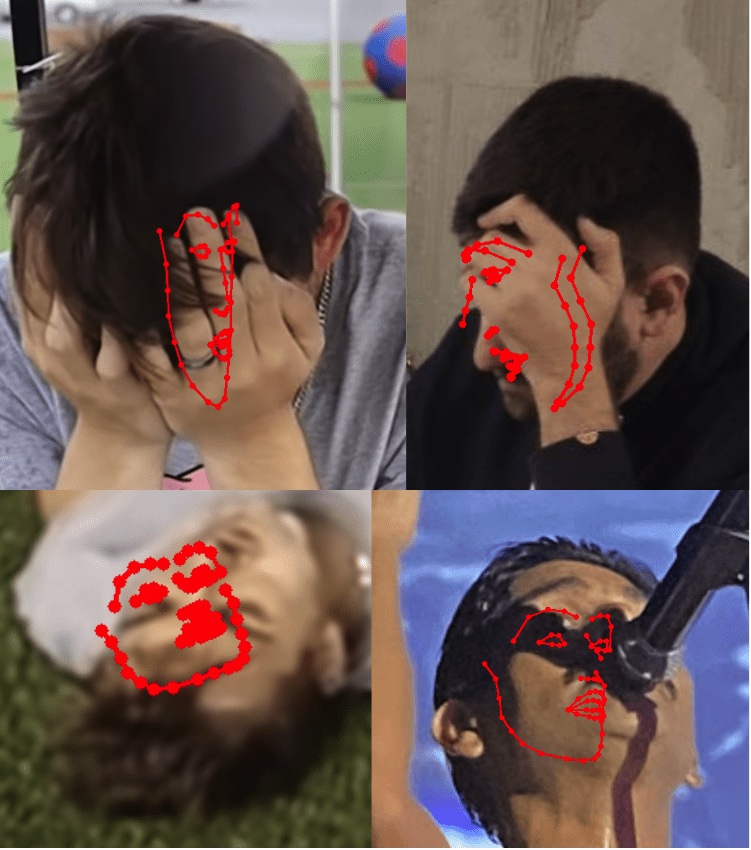}
  \caption{Failure cases of DAD-3DNet on DAD-3DHeads benchmark, 3D Landmark Localization.}
  \vspace{-1em}
  \label{fig:fails-lnmd}
\end{figure}

DAD-3DNet still sometimes fails on some cases of severe occlusions, extremely low quality, or very atypical poses (like faces upside-down) (see \cref{fig:fails-lnmd}). Our current model was trained with no augmentations, this creates a suitable venue for future explorations.

\section{Miscellaneous}

\begin{figure}[h]
\centering
\includegraphics[width=0.45\textwidth]{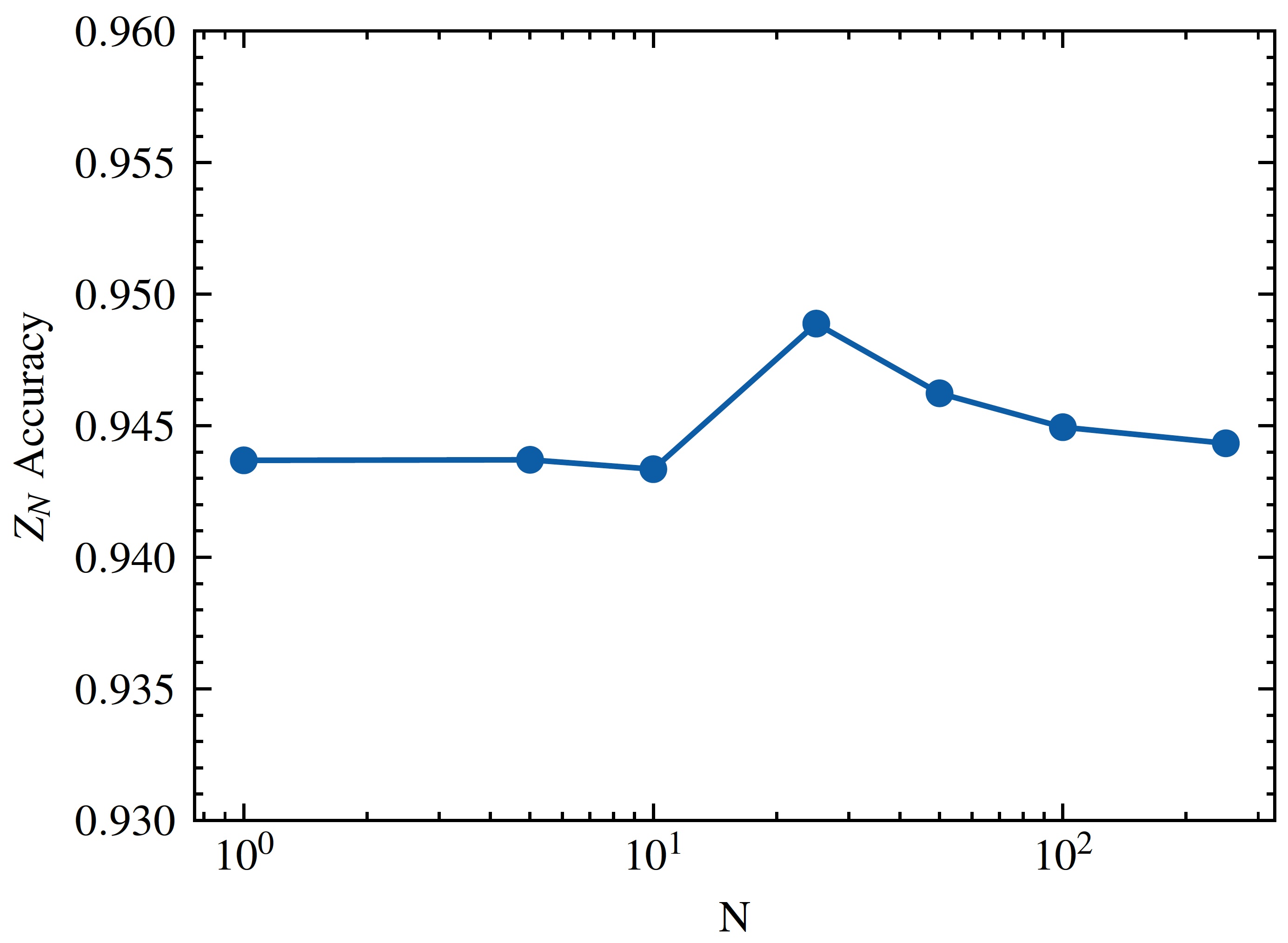}
  \caption{$Z_N$ accuracy of DAD-3DNet for different values of $n$. $x$-axis is in log-scale.}
  \vspace{-1em}
  \label{fig:z5}
\end{figure}

\textbf{\textit{Z\textsubscript{n}} Metrics.}
We measure performance of DAD-3DNet via $Z_n$ for different values of $n$, see the results in \cref{fig:z5}. The accuracy does not change dramatically with $n$, so for DAD-3DHeads benchmark we use $n=5$ as a trade-off between computational complexity and robustness.

\bigskip

\begin{figure}[h]
\centering
\includegraphics[width=0.45\textwidth]{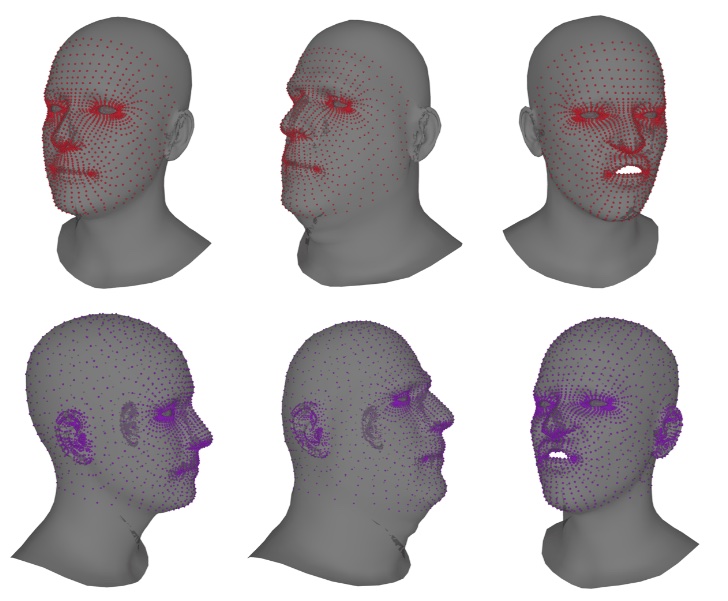}
  \caption{\textbf{"Face" and "head" subsets of FLAME mesh vertices.} Upper row: "face", capture frontal part of the head without ears. Lower row: "head", capture the head without neck. The eyeballs are excluded in both.}
  \vspace{-1em}
  \label{fig:head_and_face}
\end{figure}

\textbf{"Head" and "face" mesh vertices.} In the paper, we refer to two subsets of the FLAME mesh vertices: "head" and "face". 
The former ones are used in the loss terms calculation (see Shape+Expression loss and Reprojection loss in \cref{ssec:loss}) and in the $Z_n$ accuracy measurements for DAD-3DHeads benchmark (see \cref{ssec:benchmark}), and the latter ones - in the Chamfer distance measurements for DAD-3DHeads benchmark (see \cref{ssec:benchmark}). 
We provide visual examples of what these subsets represent on meshes with various head shape and face expressions in \cref{fig:head_and_face}.

\bigskip

\begin{figure*}[t!]
\centering
\includegraphics[width=0.9\textwidth]{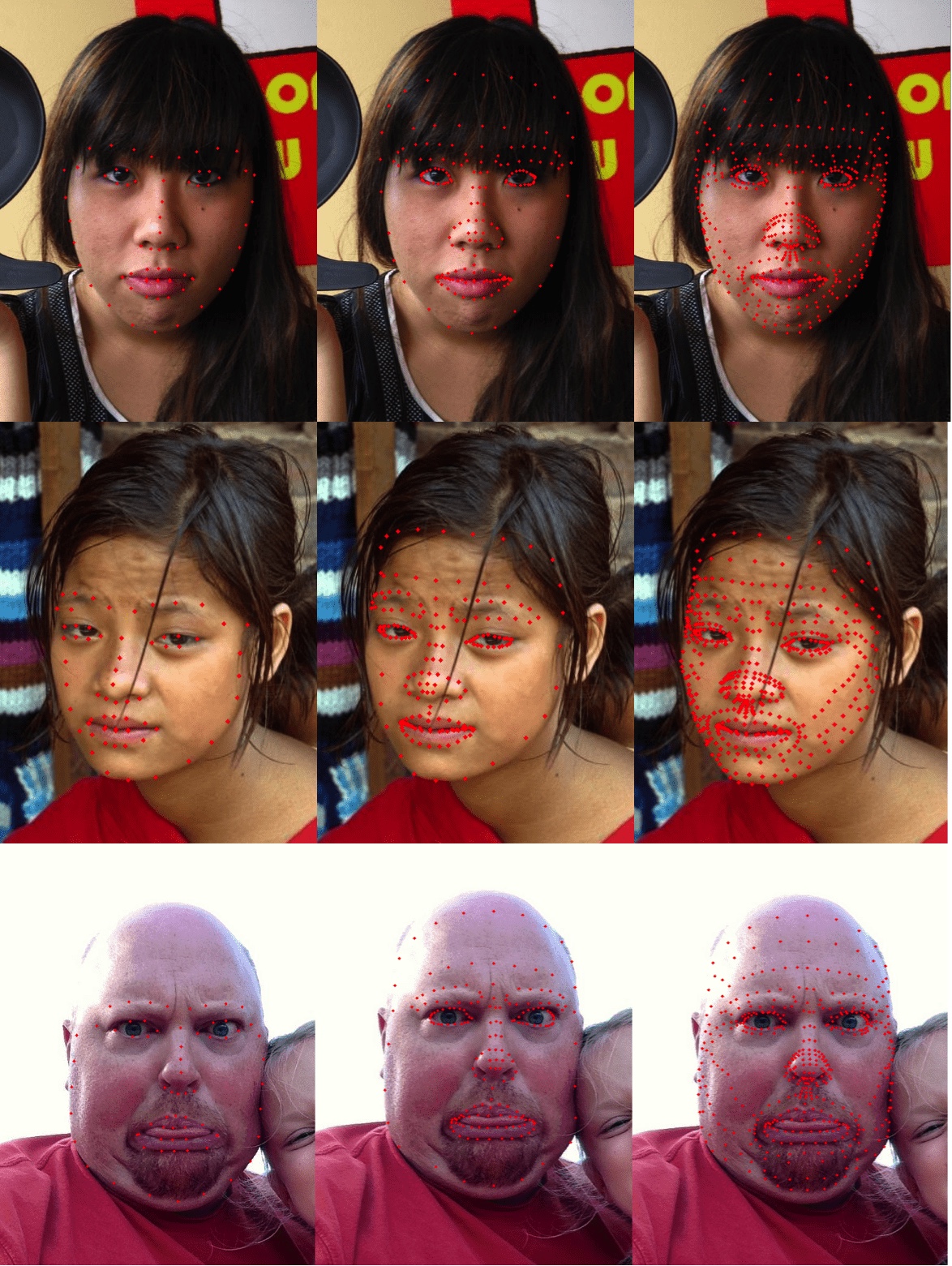}
  \caption{DAD-3DHeads dataset allows for flexibly choosing the desired landmark subset for predicting as many dense landmarks as needed. Left to right: 68 landmarks \cite{multipie}, 191 landmarks, 445 landmarks.}
  \vspace{-1em}
  \label{fig:191_landm}
\end{figure*}

\textbf{Various number of landmarks.} As DAD-3DHeads dataset is \textbf{dense}, it allows for training different models, localizing many more than the usual 68 landmarks\cite{multipie}. This flexibility saves the human annotator efforts, because the data should not be relabeled every time a different setup is needed. Moreoever, DAD-3DNet training pipeline allows for inference on any subset of head vertices, given its 3DMM prediction branch, as they can be subsampled after the entire mesh is predicted (see examples of different landmark subsets in \cref{fig:191_landm}).


\twocolumn[{%
\renewcommand\twocolumn[1][]{#1}%
\maketitle
\begin{center}
    \centering
    \captionsetup{type=figure}
  \includegraphics[width=0.85\textwidth]{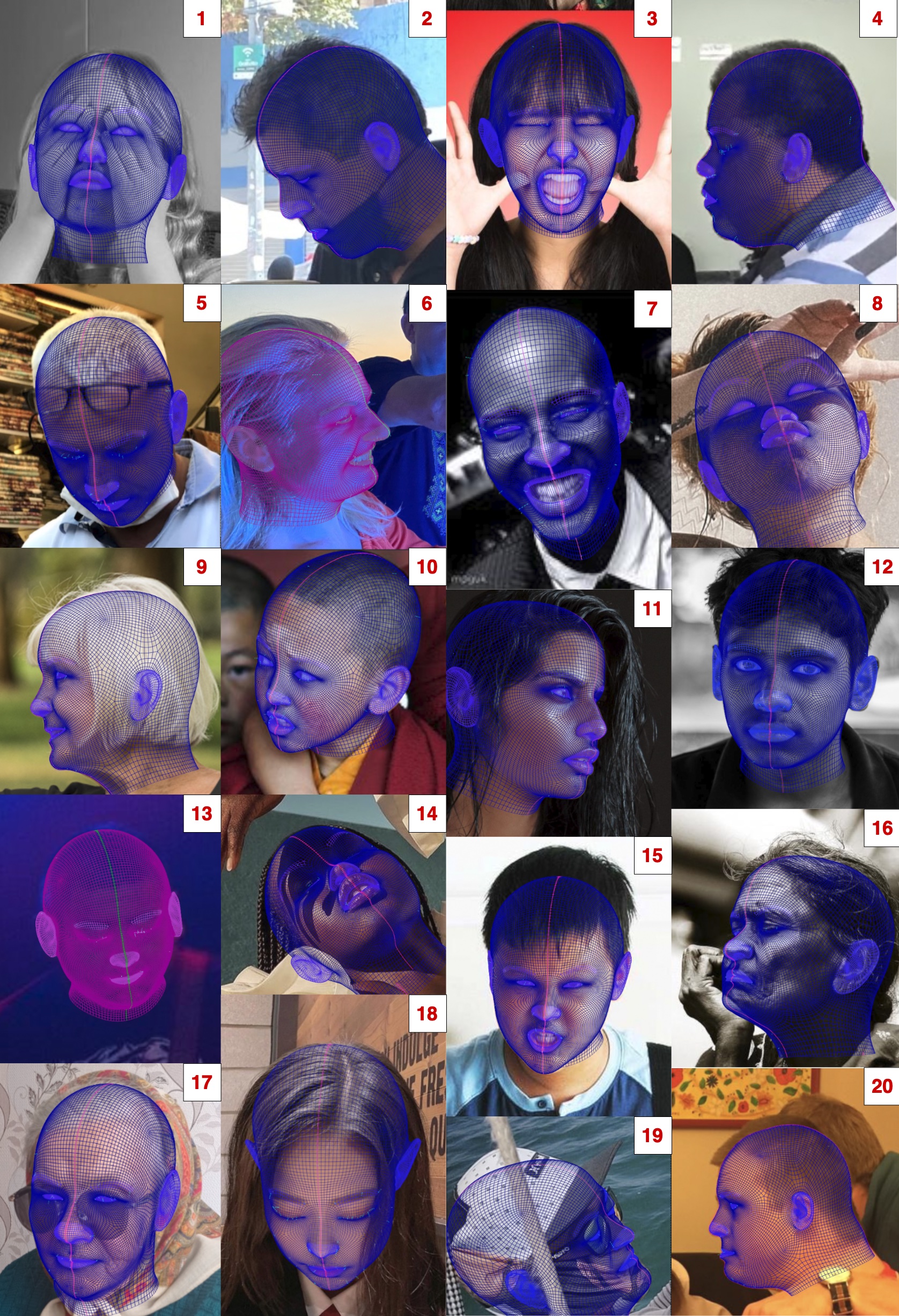}
    \caption{\textbf{DAD-3DHeads dataset, more visual examples.} The source images cover large variation in poses [2, 4, 9, 11, 14, 16, 18-20], expressions [3, 6, 7, 8, 15], occlusions [1-3, 8, 19], non-standard illumination conditions [6, 13, 14], low image quality [2, 4, 13].}
    \label{fig:diversity}
\end{center}
}]
\maketitle

\begin{figure*}[t]
\centering
    \begin{subfigure}[t]{0.85\textwidth}
        \includegraphics[width=\textwidth]{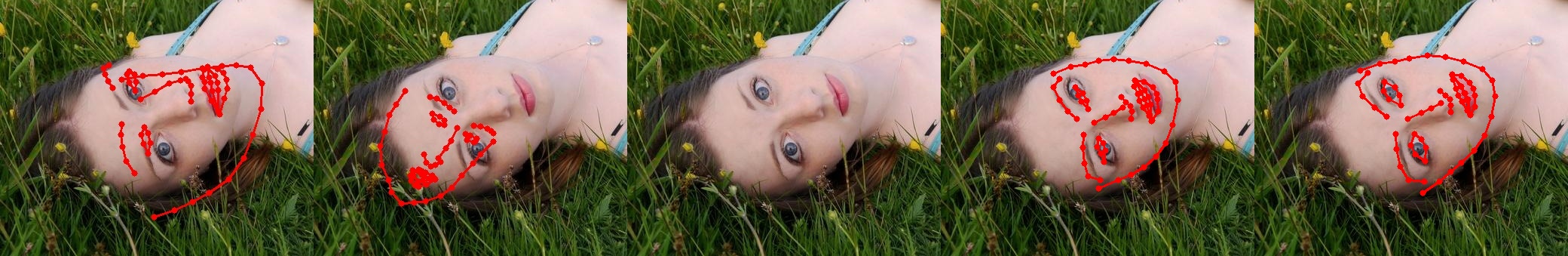}
    \end{subfigure}
    \begin{subfigure}[t]{0.85\textwidth}
        \includegraphics[width=\textwidth]{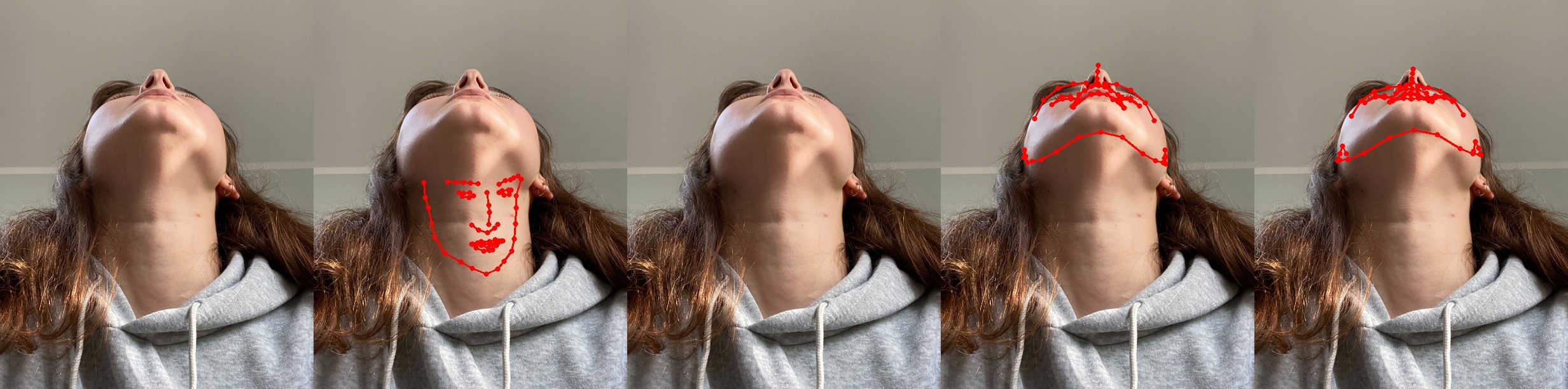}
    \end{subfigure}
    \begin{subfigure}[t]{0.85\textwidth}
        \includegraphics[width=\textwidth]{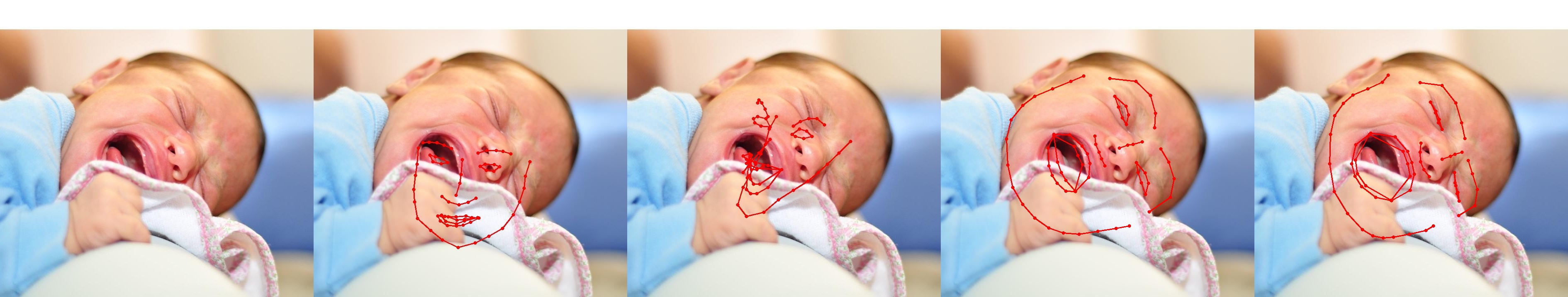}
    \end{subfigure}
    \begin{subfigure}[t]{0.85\textwidth}
        \includegraphics[width=\textwidth]{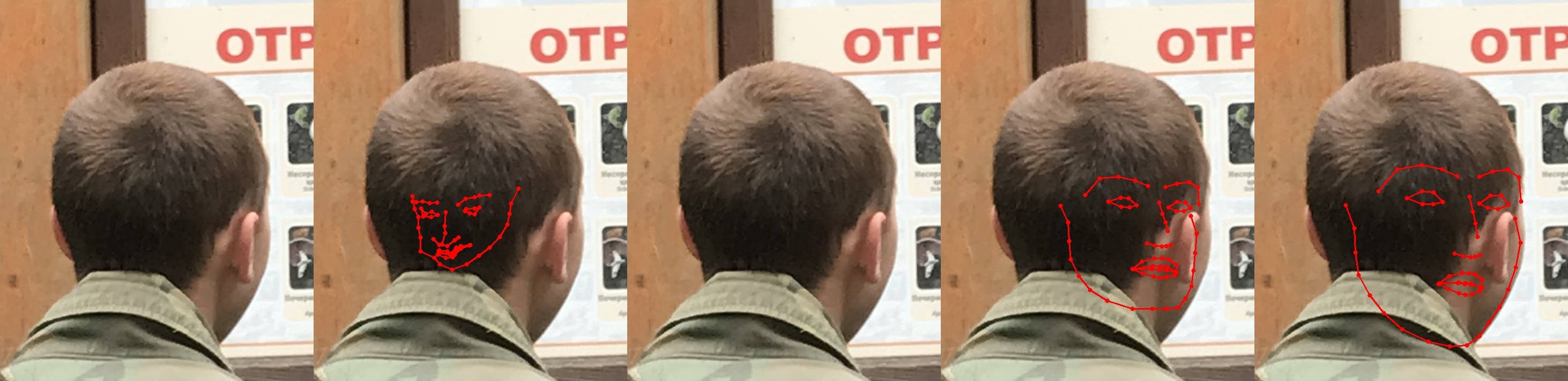}
    \end{subfigure}
        \begin{subfigure}[t]{0.85\textwidth}
        \includegraphics[width=\textwidth]{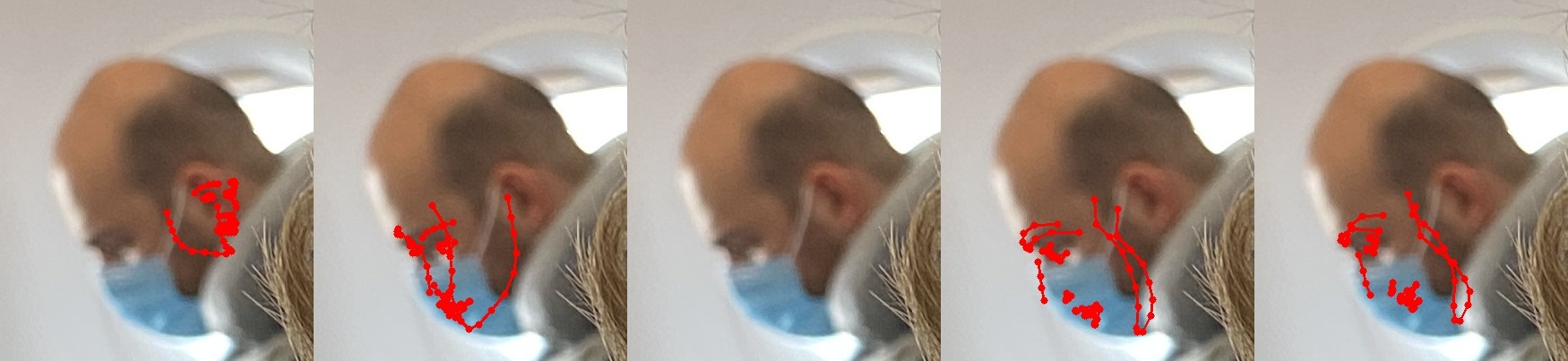}
    \end{subfigure}
        \begin{subfigure}[t]{0.85\textwidth}
        \includegraphics[width=\textwidth]{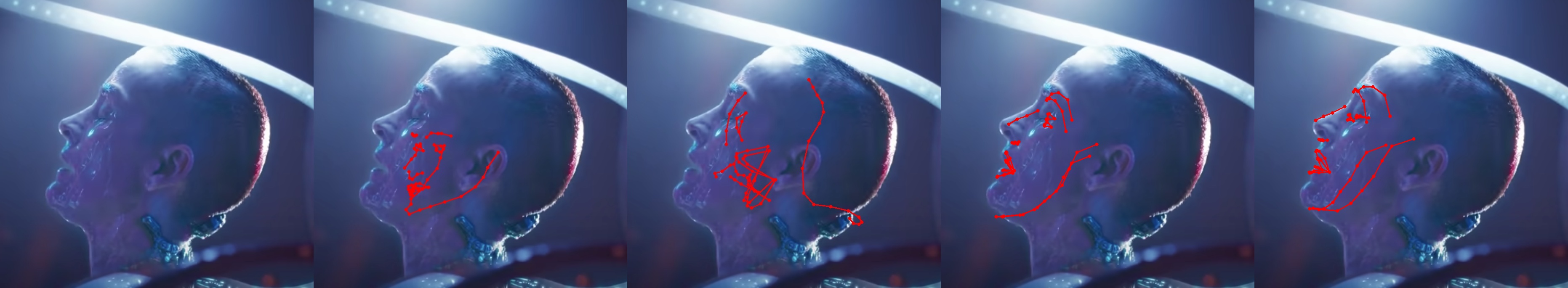}
    \end{subfigure}
\caption{Qualitative comparison of DAD-3DNet  and state-of-the-art methods on challenging cases from DAD-3DHeads benchmark. \textbf{Left to right:} 3DDFA-v2\cite{guo2020towards}, FaceSynthetics\cite{wood2021fake}, JVCR \cite{jvcr}, DAD-3DNet (ours), ground truth.}
\vspace{-1em}
\label{fig:3d_landmarks_dad_a}
\end{figure*}%
    \begin{figure*}[t]\ContinuedFloat
            \centering

        \begin{subfigure}[t]{0.85\textwidth}
        \includegraphics[width=\textwidth]{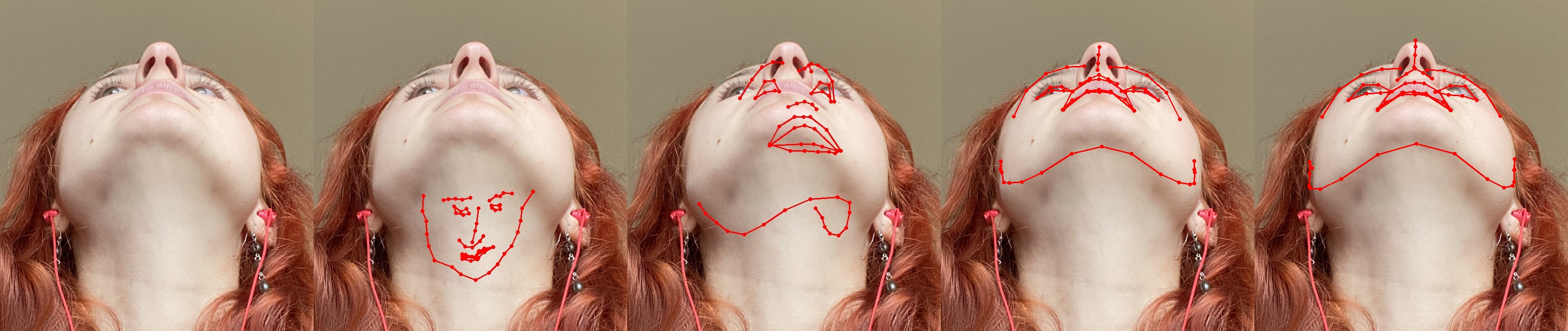}
    \end{subfigure}
        \begin{subfigure}[t]{0.85\textwidth}
        \includegraphics[width=\textwidth]{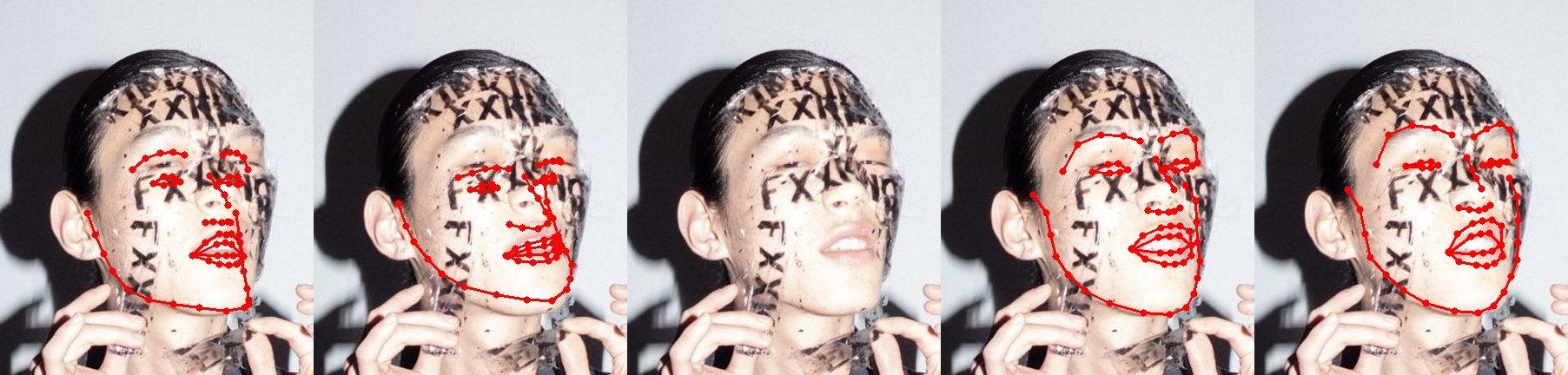}
    \end{subfigure}
        \begin{subfigure}[t]{0.85\textwidth}
        \includegraphics[width=\textwidth]{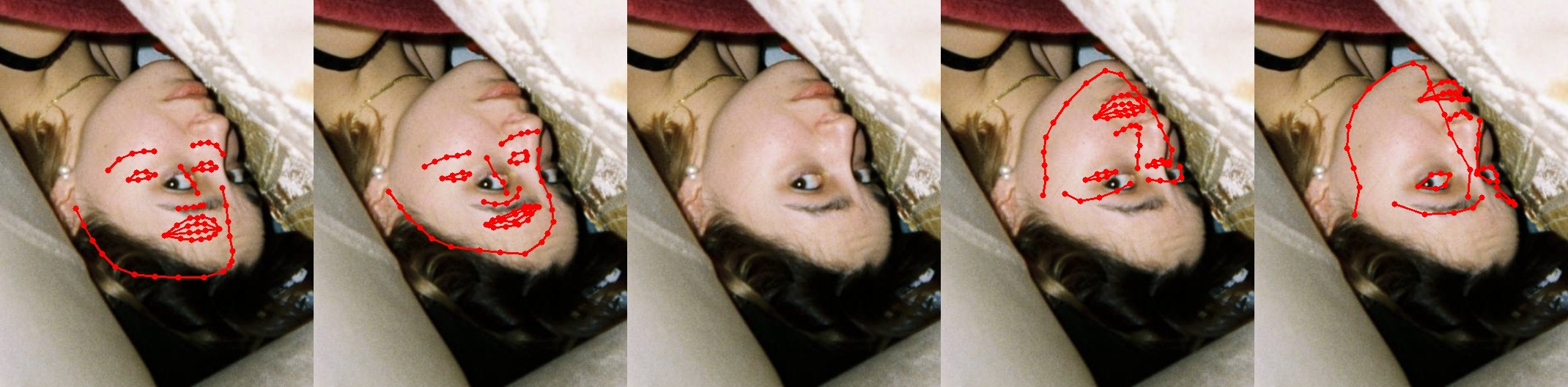}
    \end{subfigure}
    \begin{subfigure}[t]{0.85\textwidth}
        \includegraphics[width=\textwidth]{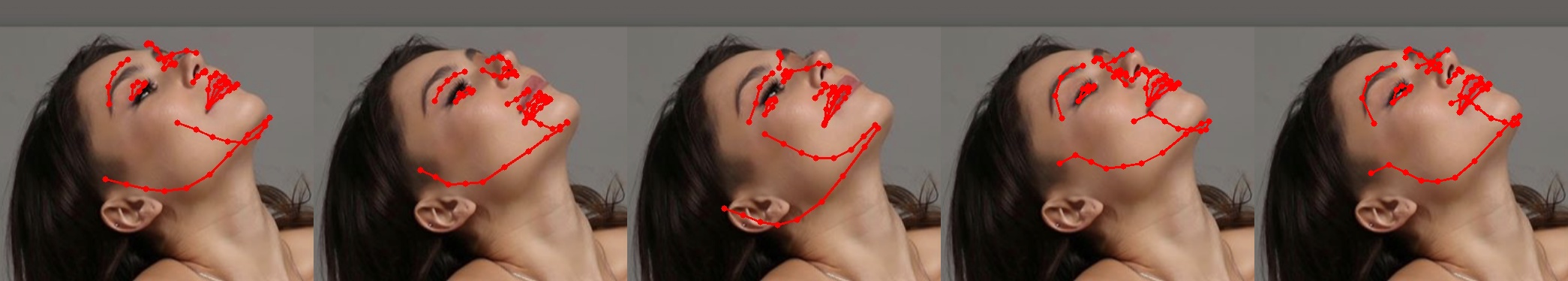}
    \end{subfigure}
        \begin{subfigure}[t]{0.85\textwidth}
        \includegraphics[width=\textwidth]{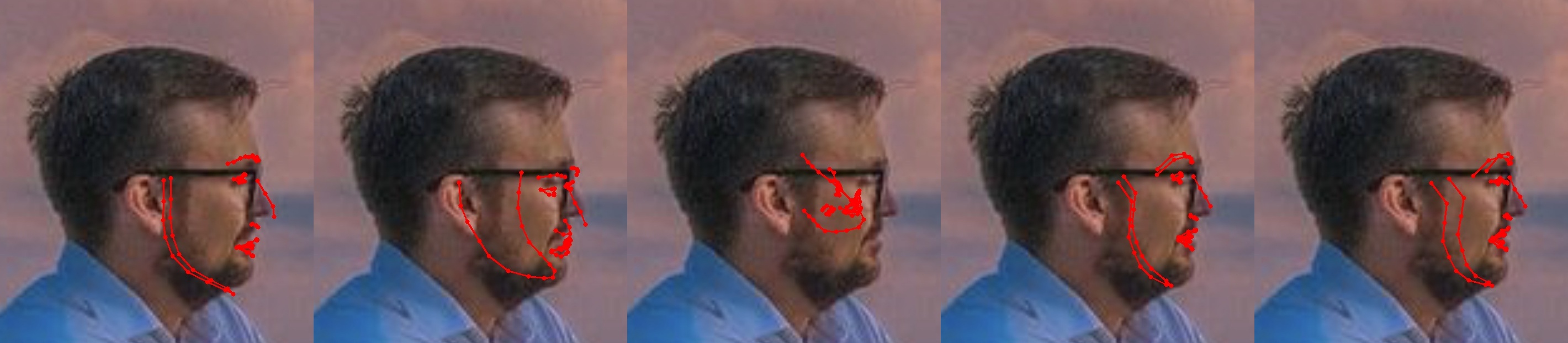}
    \end{subfigure}
            \begin{subfigure}[t]{0.85\textwidth}
        \includegraphics[width=\textwidth]{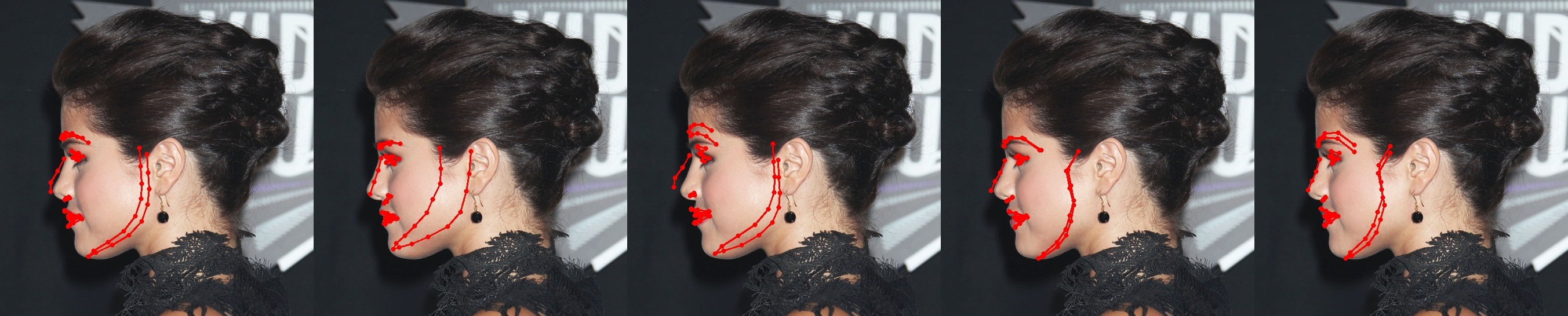}
    \end{subfigure}
  \caption{Qualitative comparison of DAD-3DNet against state-of-the-art methods on challenging cases from DAD-3DHeads benchmark (cont.)  \textbf{Left to right:} 3DDFA-v2\cite{guo2020towards}, FaceSynthetics\cite{wood2021fake}, JVCR\cite{jvcr}, DAD-3DNet (ours), ground truth.}

\end{figure*}%

    \begin{figure*}[t]\ContinuedFloat
            \centering

        \begin{subfigure}[t]{0.85\textwidth}
        \includegraphics[width=\textwidth]{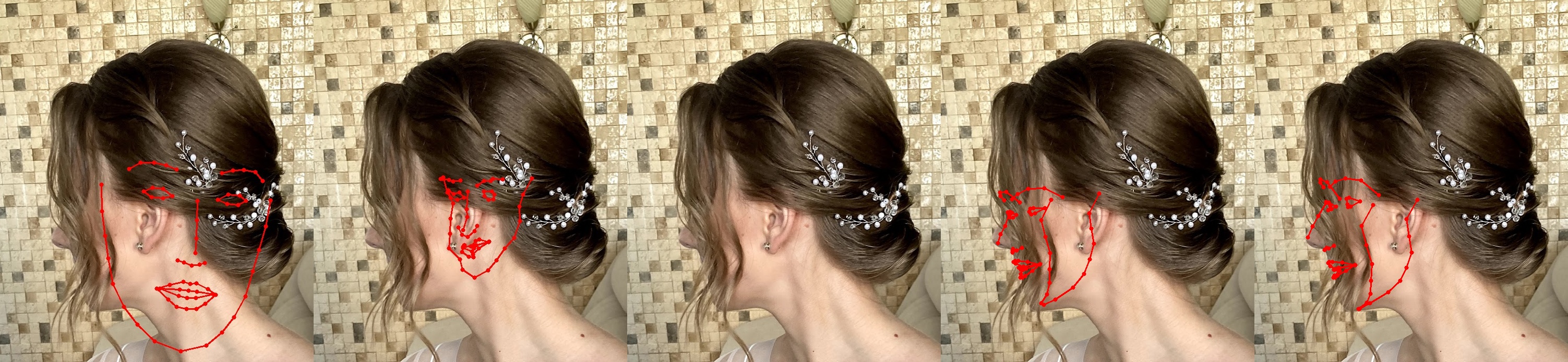}
    \end{subfigure}
        \begin{subfigure}[t]{0.85\textwidth}
        \includegraphics[width=\textwidth]{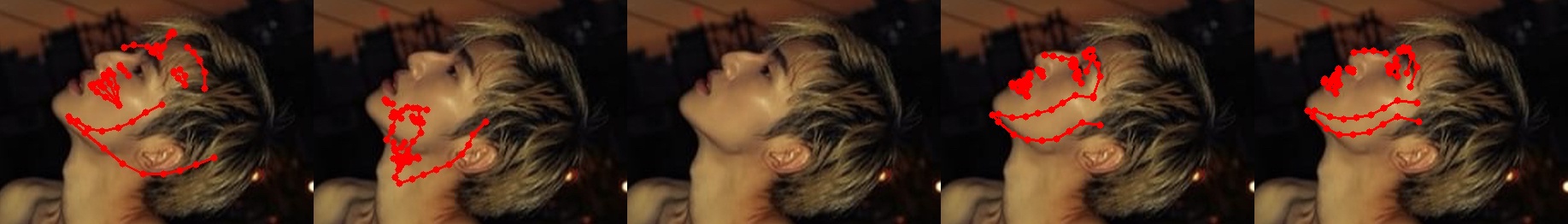}
    \end{subfigure}
        \begin{subfigure}[t]{0.85\textwidth}
        \includegraphics[width=\textwidth]{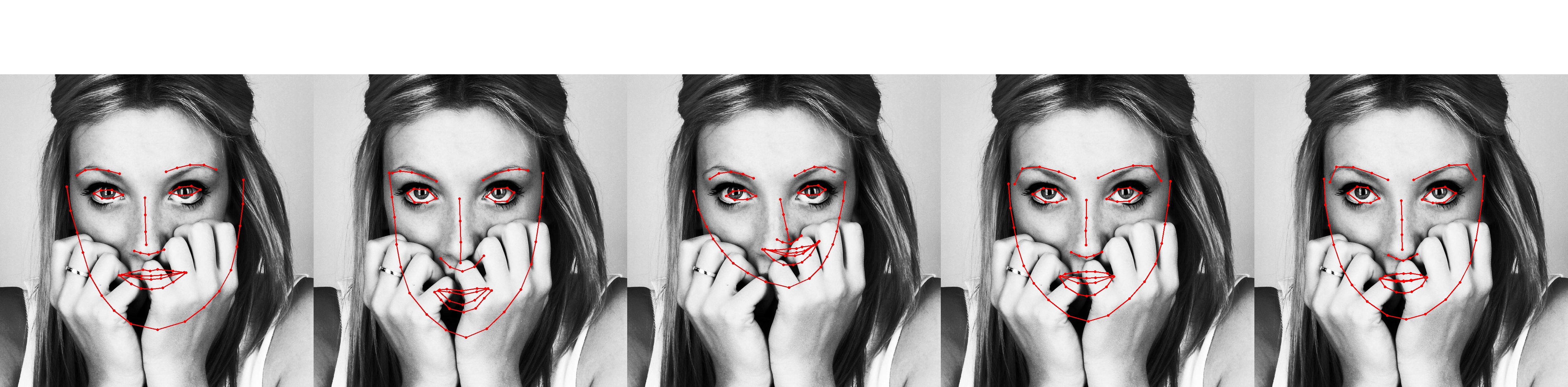}
    \end{subfigure}
    \begin{subfigure}[t]{0.85\textwidth}
        \includegraphics[width=\textwidth]{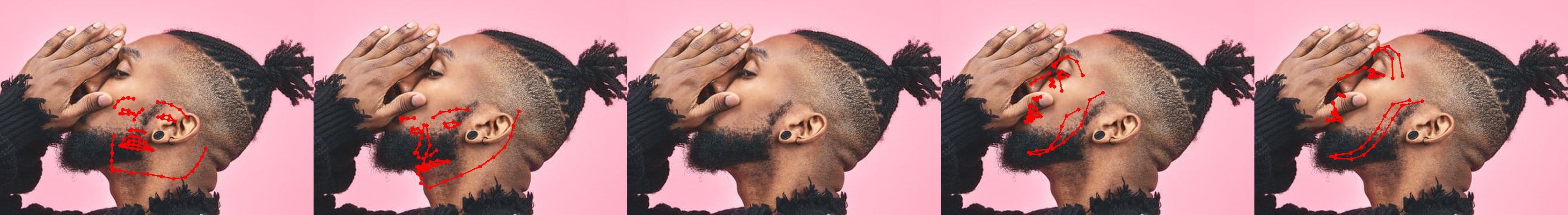}
    \end{subfigure}
        \begin{subfigure}[t]{0.85\textwidth}
        \includegraphics[width=\textwidth]{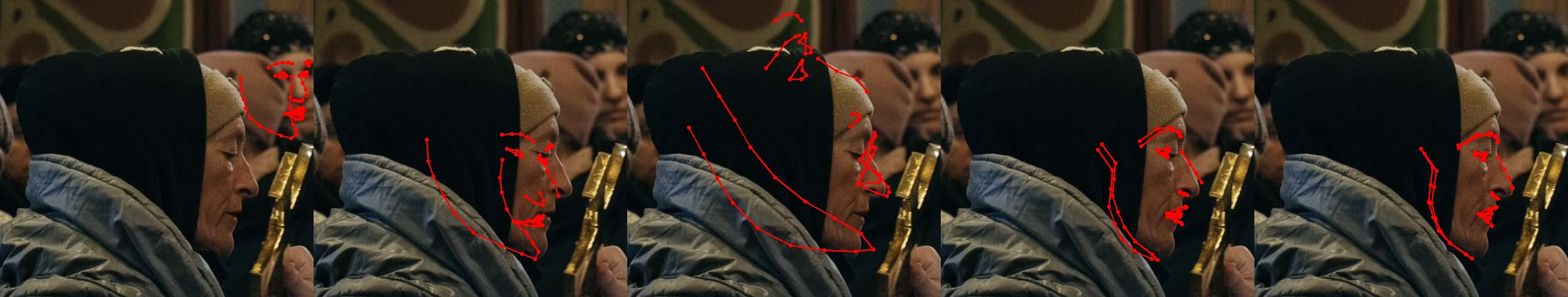}
    \end{subfigure}
            \begin{subfigure}[t]{0.85\textwidth}
        \includegraphics[width=\textwidth]{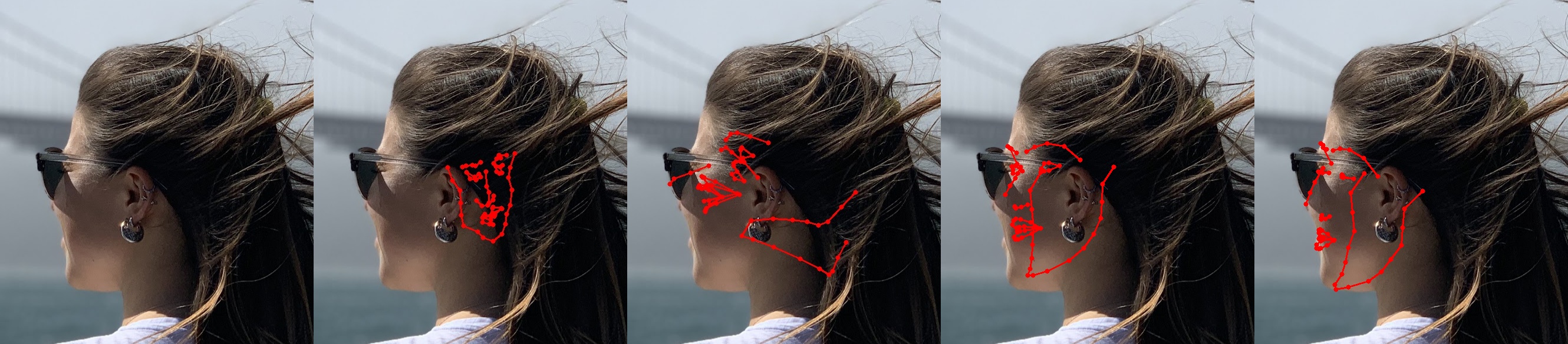}
    \end{subfigure}
  \caption{Qualitative comparison of DAD-3DNet against state-of-the-art methods on challenging cases from DAD-3DHeads benchmark (cont.)  \textbf{Left to right:} 3DDFA-v2\cite{guo2020towards}, FaceSynthetics\cite{wood2021fake}, JVCR\cite{jvcr}, DAD-3DNet (ours), ground truth.}
  \vspace{-1em}
  \label{fig:3d_landmarks_dad_c}
\end{figure*}

\clearpage

\end{document}